\pdfoutput=1
\documentclass[10pt,letterpaper]{article}

\usepackage[T1]{fontenc}
\usepackage{amsmath}
\usepackage{libertine}
\usepackage[libertine]{newtxmath}
\usepackage[varqu]{zi4}
\usepackage[letterpaper,textwidth=5.5in,textheight=9.5in,centering]{geometry}
\usepackage{microtype}
\usepackage{titlesec}
\usepackage{fancyhdr}

\usepackage{graphicx}
\usepackage{booktabs}
\usepackage{multirow}
\usepackage{array}
\usepackage{needspace}
\usepackage[table]{xcolor}
\usepackage[numbers,sort&compress]{natbib}
\usepackage{xurl}
\usepackage{float}
\usepackage[font=small,labelfont=bf]{caption}
\usepackage{tikz}
\usetikzlibrary{positioning,arrows.meta,fit,calc,decorations.pathreplacing,backgrounds}
\usepackage[colorlinks=true,allcolors=okBlue]{hyperref}

\input{glyphtounicode}
\graphicspath{{figures/}}

\fancypagestyle{plain}{\fancyhf{}\fancyfoot[C]{\footnotesize\thepage}}

\titleformat{\section}[hang]{\large\bfseries\color{okInk}}{\thesection}{0.6em}{}[{\color{okBlue}\titlerule[0.6pt]}]
\titleformat{\subsection}[hang]{\normalsize\bfseries\color{okInk}}{\thesubsection}{0.5em}{}
\titleformat{\paragraph}[runin]{\bfseries\color{okBlue}}{}{0pt}{}
\titlespacing*{\section}{0pt}{2.2ex plus 0.6ex}{1.1ex}
\titlespacing*{\subsection}{0pt}{1.8ex plus 0.5ex}{0.7ex}
\titlespacing*{\paragraph}{0pt}{1.2ex plus 0.3ex}{0.6em}

\definecolor{okBlue}{HTML}{0072B2}
\definecolor{okSky}{HTML}{56B4E9}
\definecolor{okGreen}{HTML}{009E73}
\definecolor{okOrange}{HTML}{E69F00}
\definecolor{okVermillion}{HTML}{D55E00}
\definecolor{okPurple}{HTML}{CC79A7}
\definecolor{okGray}{HTML}{7F7F7F}
\definecolor{okLight}{HTML}{C8C8C8}
\definecolor{okInk}{HTML}{1A1A1A}

\tikzset{
  every picture/.style={font=\scriptsize, color=okInk,
                        node distance=3pt and 8pt},
  box/.style={draw, rounded corners=1.5pt, inner sep=2.5pt, align=center,
              minimum height=11pt},
  mem/.style={box, fill=black!7},
  cell/.style={draw, inner sep=0pt, minimum height=9pt, font=\tiny,
               anchor=north west},
  flow/.style={-{Stealth[length=4pt]}, semithick},
  title/.style={font=\scriptsize\bfseries, align=center},
  note/.style={font=\scriptsize\itshape, align=center},
  panel/.style={draw, black!35, rounded corners=3pt, inner sep=4pt},
}

\newcommand{\bw}{b/weight}
\newcommand{\Planes}{\textsc{Planes14}}
\newcommand{\Tetra}{\textsc{Tetra}}
\hypersetup{
  pdftitle={Tetra: Serving Leech-Lattice Quantized LLMs at 2.7 Bits per Parameter},
  pdfauthor={Pier-Jean Malandrino}}

\newcommand{\titleblock}{%
  \begingroup\centering
  {\color{okBlue}\rule{\textwidth}{1.2pt}}\\[1.0ex]
  {\LARGE\bfseries Tetra: Serving Leech-Lattice Quantized LLMs\\[2pt]
   at 2.7 Bits per Parameter\par}
  \vspace{1.0ex}
  {\color{okBlue}\rule{\textwidth}{0.5pt}}\\[1.3ex]
  {Pier-Jean Malandrino\par}
  {\footnotesize Scub, Bordeaux, France \;\textperiodcentered\; \texttt{pierjean.malandrino@scub.net}\par}
  \vspace{0.5ex}
  {\footnotesize\textsc{November 2026} \;\textperiodcentered\; preprint, not peer reviewed\par}
  {\footnotesize Code, measurement logs and preregistrations: \url{https://github.com/pjmalandrino/llvq}\par}
  \vspace{1.6ex}\endgroup}

\begin{document}
\titleblock\thispagestyle{plain}

\begin{abstract}
\noindent
Leech-lattice quantization gives good quality at two bits per weight, but its
codebooks hold more than $10^{14}$ points, too many for a lookup table. Our
earlier kernel expanded the codes at load time and read 4.804 bits per weight
from GPU memory for 2 bits of code. We present \Tetra{}, a new codebook on the
same lattice. A 24-weight block still takes 48 bits, most of which index a
64-state trellis of the Golay code and one shared 16~KiB table. The kernel
decodes a block with six table loads and two small lookups inside the
matrix-vector product, and reads 2.150 bits per weight. For full models, we
retrain one scale per matrix row, store the matrices that lose the most as
4-bit integers, and pay for them with 4-bit embedding tables. Our Qwen3-4B,
8B and 14B files hold 2.73, 2.70 and 2.73 bits per parameter over the whole
model. They score 63.37, 69.58 and 75.66 on the full MMLU test set, 4.76,
4.21 and 2.46 points below 4-bit AWQ at 5.3 to 6.0 bits per parameter, and
their WikiText-2 perplexity is 2.8 to 8.0\,\% above FP16's. On GSM8K, through
the kernel, they lose 9.63, 4.62 and 3.26 points to FP16. A persistent kernel
runs each token in one launch and returns the same tokens as one launch per
matrix. With it the files generate 221.5, 142.5 and 86.3 tokens per second on one
NVIDIA L40S. Against llama.cpp's IQ2\_S at the same bits per parameter, on the
same card, our files score 9.08, 4.13 and 2.89 MMLU points higher. llama.cpp
decodes 33.5 to 63.9 tokens per second faster and holds less card memory after
a long prompt. On an A100, launched once per matrix, our files decode slower
than FP16 in vLLM. We preregistered the main experiments.
\end{abstract}

\section{Introduction}
\label{sec:intro}

Qwen3-14B needs 29.5~GB for its FP16 weights alone, more than a 24~GB
consumer GPU holds. At batch 1 a model reads every weight once per
token, so weight size sets both memory and speed. Two or three bits per
parameter let such a model run on local hardware.

\begin{figure}[t]
\centering
\includegraphics{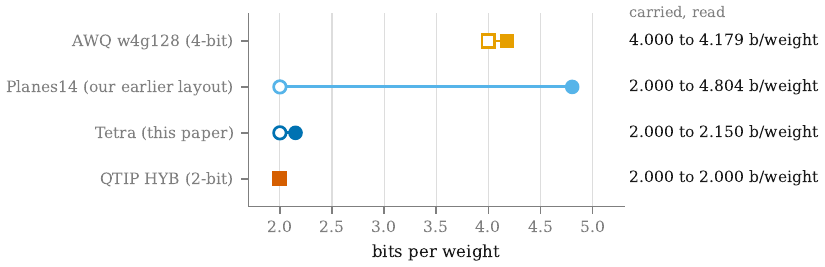}
\caption{Bits of code a format carries per weight (hollow marker) and bits its
kernel reads per weight in VRAM (filled marker). Circles are our formats,
squares deployed kernels. \Planes{}, our earlier layout, reads $2.40\times$ its 2.000
bits of code as it expands its codebook first, and \Tetra{} $1.07\times$.
\Tetra{}'s extra bits are its row scales and the unquantized tail of each row,
both in f32 in this benchmark (\S\ref{sec:file}). QTIP's marker counts its
codes alone (Appendix~\ref{app:bench}). FP16
(16.000~\bw{}) is left out. Data: echelle-formats.csv.}
\label{fig:gap}
\end{figure}

Van der Ouderaa et al.~\citep{llvq2026} quantize blocks of 24 weights to points of the Leech lattice
$\Lambda_{24}$. At two bits per weight, their method gives the best quality in
their comparison, ahead of QuIP\#~\citep{quipsharp2024} and
QTIP~\citep{qtip2024}. Each block becomes one 48-bit code, and the codebook is
far too large to store (\S\ref{sec:background}).
In earlier work~\citep{llvq1preprint} we expanded each code at load time into
a wider format decoded with shifts and masks. The kernel then read 4.804 bits
per weight from GPU memory (VRAM) for 2 bits of code, more than the 4.179 of
4-bit AWQ, and the saving was lost (Figure~\ref{fig:gap}).

This paper removes that loss with \Tetra{}, a new codebook on the same
lattice, then builds complete models around it. The lattice is built from the
Golay code, a binary code of length 24. Three of its codewords with eight ones
cover all 24 coordinates. Read in those three groups of eight, the code is a
trellis~\citep{forney1988}: a layered graph with 64 nodes at each cut, where
each codeword is one path. A block is stored as its state, a few bits that
choose the path and the gain, and three 11-bit indices into one 16~KiB table,
still 48 bits. The kernel reads 2.150 bits per weight.

Our contributions:
\begin{itemize}
\item \Tetra{}, the rule it imposes on which lattice points to keep, and the
  measured cost of that rule (\S\ref{sec:tetra}).
\item A fused kernel that decodes and multiplies in one pass, timed against
  FP16, AWQ~\citep{awq2024} and our earlier layout on the same 252 matrices,
  on two GPUs (\S\ref{sec:kernel}, \S\ref{sec:a100}).
\item A persistent kernel that runs a whole token in one launch and returns
  the tokens of the per-matrix path. It takes the 4B from 113.8 to 221.5
  tokens per second (\S\ref{sec:4e}).
\item A recipe for complete files at about 2.7 bits per parameter that changes
  no lattice code: retrained row scales, int4 for the projections that lose the
  most, and 4-bit embedding tables to pay for them. On Qwen3-4B, the lattice
  file with int4 value matrices scores about 12 MMLU points below FP16, and the
  recipe closes part of that gap. Each step is measured at 4B, 8B and 14B with
  paired intervals over the full MMLU test set (\S\ref{sec:model},
  \S\ref{sec:chain}).
\item MMLU, perplexity, GSM8K through the kernel, speed and memory of our files
  against FP16 and AWQ at three sizes (\S\ref{sec:main}, \S\ref{sec:gsm8k}).
  With about as much weight memory or less, our 8B and 14B score above AWQ's
  4B and 8B on MMLU and are not separated from them on GSM8K
  (\S\ref{sec:sizeup}).
\item A preregistered comparison with llama.cpp's IQ2\_S at the same bits per
  parameter, at three sizes on the same card: quality, speed and card memory
  (\S\ref{sec:duel}), then speed, time to the first token and peak memory
  after prompts of up to 8{,}192 tokens (\S\ref{sec:context}).
\end{itemize}

These files trade quality for memory. They use about half of AWQ's bits per
parameter, or about 65\,\% if AWQ also stored its embedding tables in 4 bits
(\S\ref{sec:main}). In return they do not match AWQ on MMLU or GSM8K, although
their MMLU gap to AWQ at 14B is about half the gap at 4B. On GSM8K, run
through the kernel, they lose more to FP16 than on MMLU when counted in
errors, at every size (\S\ref{sec:gsm8k}). At the same bits, they score above
llama.cpp's IQ2\_S on MMLU and perplexity at every size, and llama.cpp
decodes faster (\S\ref{sec:duel}). For the experiments behind the main
results, we fixed the plan and our predictions before measuring, and we report
the predictions that were wrong (Appendix~\ref{app:record}).

\section{Background and related work}
\label{sec:background}

\paragraph{Weight-only quantization at batch 1.} GPTQ~\citep{gptq2023}
quantizes a matrix one column at a time and pushes each rounding error onto the
columns not yet quantized, using second-order statistics from calibration text.
AWQ~\citep{awq2024} rescales input channels by activation size before rounding
to 4 bits. With groups of 128 weights it is the usual 4-bit baseline, which
vLLM serves with the Marlin kernel~\citep{vllm2023,marlin2024}. Two recent
methods improve on GPTQ. Qronos~\citep{qronos2025} also corrects the error
that quantizing the previous layers leaves in a layer's inputs.
KronQ~\citep{kronq2026} models the Hessian with Kronecker factors that include
gradient information, uses the output-side factor in its incoherence
processing, and allocates precision across layers from a sensitivity measure.
Our encoder does none of this, and our int4 choices were measured on a
benchmark instead (\S\ref{sec:chain}).

\paragraph{Two-bit vector quantization.} Below three bits, the strong methods
encode several weights at once, as a choice from a codebook.
QuIP\#~\citep{quipsharp2024} uses the $E_8$ lattice. QTIP~\citep{qtip2024}
uses a trellis. Its hybrid code hashes each state to index a 2~KiB table, and
its computed codes, such as 3INST, need no table. AQLM and
VPTQ~\citep{aqlm2024,vptq2024} learn codebooks of at most $2^{16}$ entries and
decode by reading them, and both papers tie kernel speed to how well these
tables fit in the GPU cache. The IQ2 formats of llama.cpp~\citep{llamacpp}
decode from grids of 256 to 1{,}024 points held in a lookup table.
GLVQ~\citep{glvq2025} learns a lattice for each group of weights, through a
learned generator matrix, and decodes it with a matrix-vector product and an
element-wise inverse $\mu$-law. At 2 bits it reports a lower WikiText-2
perplexity than QTIP and QuIP\# on Llama~2-70B. Its 32-dimensional variant
decodes Llama~2-7B at 82.0 tokens per second on one RTX 4090, against 105.2
for QTIP. LiftQuant~\citep{liftquant2026}
projects a 1-bit lattice from a higher-dimensional space, which makes its bit
width nearly continuous. Both learn the codebook to gain quality. We keep the
codebook of \citet{llvq2026} and change how the GPU reads it.

\paragraph{The Leech lattice.} $\Lambda_{24}$ is the densest sphere packing
in 24 dimensions~\citep{cohn2017}. Inside a GPTQ-style loop,
\citet{llvq2026} code each block as a lattice direction (the shape) and a
length (the gain). Their best codebook on Gaussian data spends 47 bits on one of
$1.1\times10^{14}$ lattice points inside a ball and one bit on the gain. The
model results we cite spend all 48 bits on a larger ball of
$2.8\times10^{14}$ points. On Qwen3-4B they report 60.7 on MMLU without
fine-tuning and 62.8 with it. Their GPU kernel decodes only one shell of the
lattice (the points at one distance from the origin), while those scores use
several shells.

\paragraph{Our earlier kernel.} In \citet{llvq1preprint} we served their
47-bit codebook with its gain bit. At load time, each code was unfolded into a
112-bit \Planes{} record: a class, a gain bit, a sign mask and three bit planes
(Figure~\ref{fig:word}).

\section{The Tetra codebook}
\label{sec:tetra}

\subsection{The lattice and the quantizer}

We use the integer form of the Leech lattice, up to the overall scale that the
quantizer sets. A vector of 24 integers $y\in\mathbb{Z}^{24}$ lies in
$\Lambda_{24}$ when it can be written as
\begin{equation}
y_j = p + 2c_j + 4k_j,\qquad \textstyle\sum_j k_j \equiv p \pmod 2 .
\label{eq:leech}
\end{equation}
Here $p$ is a bit, the $k_j$ are integers, and $c$ is a codeword of the binary
Golay code $\mathcal{G}_{24}$. Two global constraints pick the lattice out of
$\mathbb{Z}^{24}$: $c$ is one of the $2^{12}$ codewords among the $2^{24}$ bit
patterns, and one parity check ties all 24 coordinates. A decoder has to
respect both.

The encoder uses the shape-gain scheme of \citet{llvq2026}. It applies the
input-side rotation of incoherence
processing~\citep{quip2023,quipsharp2024,quarot2024}, and runs a GPTQ-style
loop~\citep{gptq2023}. For each block of 24 rotated weights, it looks for the lattice point under a
length cap whose direction is closest to the block, and rebuilds the weights as
\begin{equation}
\hat{w}_j = \frac{y_j}{\lVert y\rVert}\,\mathrm{gain}[g]\,\sigma_{\mathrm{row}},
\qquad \lVert y\rVert=\sqrt{16m},
\label{eq:recon}
\end{equation}
where $m$ is the shell index, $g$ selects the gain, and $\sigma_{\mathrm{row}}$
is one scale per output row, stored in f64 and read by the kernel in f32. We
change only the region the encoder searches (\S\ref{sec:kept}) and the bits it
writes for each block. The \Tetra{} search is not exhaustive. At two trial
scales, the encoder keeps a few candidates per section and joins the best over
the trellis. It runs once, offline, on a CPU.

\subsection{Three octads and a trellis}
\label{sec:octads}

An octad is a Golay codeword with exactly eight ones. We fix three octads that
share no position and together cover all 24 coordinates, with supports
\texttt{0x00149f}, \texttt{0x0f6840} and \texttt{0xf08320} (bit $i$ is
coordinate $i$), and read the coordinates octad by octad, each in increasing
order. Cutting after coordinates 8 and 16 splits the block
into three sections, one per octad. These cuts turn the code into the trellis
of \citet{forney1988}, a layered graph whose paths spell out bit patterns
(Figure~\ref{fig:geom}). The node a path crosses at a cut is its state. Each
cut has 64 states, and no split of $\mathcal{G}_{24}$ into three sections of
eight has fewer. The trellis has $64\times2\times16\times2 = 4096 =
|\mathcal{G}_{24}|$ paths, one per codeword, which we check when we build the
tables. Forney uses such trellises for efficient maximum-likelihood decoding.
We use this one to give every entry of a quantizer codebook an address.

\begin{figure}[t]
\centering
\begin{tikzpicture}[x=1cm, y=1cm]
\node[box, font=\scriptsize] (s0) at (0.5,1.3) {$1$};
\node[mem, font=\scriptsize, minimum width=9mm, minimum height=13mm]
  (s8) at (4.2,1.3) {$64$};
\node[mem, font=\scriptsize, minimum width=9mm, minimum height=13mm]
  (s16) at (8.2,1.3) {$64$};
\node[box, font=\scriptsize] (se) at (11.9,1.3) {$1$};
\draw[flow] (s0) -- node[above, font=\scriptsize]{2 prefixes}
                   node[below, font=\scriptsize]{$b_1$: 1 bit} (s8);
\draw[flow] (s8) -- node[above, font=\scriptsize]{16 branches}
                    node[below, font=\scriptsize]{$b_2$: 4 bits} (s16);
\draw[flow] (s16) -- node[above, font=\scriptsize]{2 suffixes}
                     node[below, font=\scriptsize]{$b_3$: 1 bit} (se);
\node[font=\scriptsize, anchor=north, align=center] at (4.2,0.45)
  {$s_8$: the 6 bits\\the word stores};
\node[font=\scriptsize, anchor=north, align=center] at (8.2,0.45)
  {$s_{16}$: not stored,\\returned with the byte};
\node[font=\scriptsize, anchor=north] at (6.2,-0.55)
  {$64\times2\times16\times2 \;=\; 4096 \;=\; |\mathcal{G}_{24}|$};
\end{tikzpicture}
\caption{The Golay code in octad order is a trellis, and a codeword is a path
through it.}
\label{fig:geom}
\end{figure}

\subsection{State, rank vectors, and one table for three sections}
\label{sec:state}

The 48-bit word of a block (\S\ref{sec:decode}) holds eight bits of state: the
trellis state $s_8$ (6 bits), the shared parity $p$, and $r$, the parity of
$\sum k_j$ over section 1. These are all that the rest of the block needs from
section 1, so a cut has $64\times4 = 256$ lattice states.

Each section has a $k$-parity, the parity of the sum of its $k_j$. Only
section 1's is stored, as $r$. Section 2's, $\delta$, is read from its row
index (explained below). Section 3's is
\begin{equation}
r_3 = p \oplus r \oplus \delta ,
\label{eq:xor}
\end{equation}
so the parity constraint of Equation~\eqref{eq:leech} costs one XOR and no bits.

A section's pattern byte is the eight bits of $c$ that fall in it. Once $p$
and this byte are known, coordinate $j$ can only take the values
$o_j + 4\mathbb{Z}$, with $o_j = p + 2c_j$. We store its \emph{rank}, its
position in that list counted outward from zero, positive first on a tie,
instead of its value. For $o = 0, 1, 2, 3$ the lists start $0, +4, -4$;
$+1, -3, +5$; $+2, -2, +6$; and $-1, +3, -5$. A section is then a rank vector
$\rho \in \{0,\dots,7\}^8$, which fits in one 32-bit table row, rank $j$ in
bits $4j$ to $4j+3$.

Two properties let a single table serve every section of every block. First,
sorting rank vectors by $\mathrm{cost}(\rho)=\sum_j(2\rho_j+1)^2$, a stand-in
for squared distance, gives the same order for every $p$ and every pattern
byte. ``Keep the 2{,}048 cheapest'', ties broken by $\rho$ in lexicographic
order, then means one list instead of 256, one
for each parity and each of the 128 pattern bytes a section admits. Second,
$\mathrm{cls}(\rho)=\sum_j[\rho_j\in\{1,2,5,6\}] \bmod 2$ equals the parity
of $\sum_j k_j$ for every pattern. We sort the table by this class, so a row's
position gives its parity with no extra field.

The table holds 4{,}096 rows: the 2{,}048 cheapest rank vectors of class 0,
then the 2{,}048 cheapest of class 1. Eleven bits pick a row within one half.
The middle section reads the same rows through its own list of the 2{,}048
cheapest overall: $N_0 = 1240$ rows of class 0, then 808 of class 1. Its index
alone thus gives the class of its row, the $\delta$ that
Equation~\eqref{eq:xor} needs.

\subsection{The word and the decode}
\label{sec:decode}

\begin{figure}[t]
\centering
\includegraphics{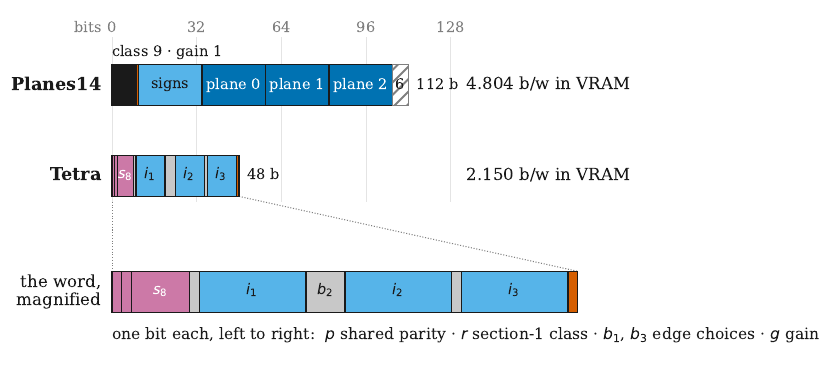}
\caption{The \Planes{} record of our earlier layout and the \Tetra{} word,
drawn to the same scale in bits, with the word magnified below to label its
ten fields. \Planes{} unfolds the 47-bit index and its gain bit into 112 bits,
one record every 14 bytes. The file stores the \Tetra{} word and the kernel
reads it unchanged, 48 bits, one word every 6 bytes. Purple marks the eight bits of state,
grey the edge choices, blue the row indices into the shared table and red the
gain bit. Hatching is padding.}
\label{fig:word}
\end{figure}

Figure~\ref{fig:word} shows the resulting 48-bit word next to the \Planes{}
record it replaces. Each field can take a power-of-two number of values, so
any 48-bit value is a legal word. A benchmark can therefore stream random
words and check every decode against a reference. Figure~\ref{fig:decode}
shows the decode.

\begin{figure}[t]
\centering
\begin{tikzpicture}[x=1cm, y=1cm,
  load/.style={box, fill=white, minimum width=3.7cm}]
\node[font=\footnotesize\bfseries] at (5.4,4.55)
  {one block: six loads, one waits for another};
\node[box, minimum width=3cm] (w) at (5.4,3.95) {the 48-bit word};
% the trellis tables: the only dependent read is the suffix
\node (Lt) at (2.5,2.85) {\textbf{trellis tables}, 2{,}304~B};
\node[load] (P) at (2.5,2.3) {\texttt{prefixes}$[2s_8{+}b_1]$};
\node[load] (B) at (2.5,1.45) {\texttt{branches}$[16s_8{+}b_2]$};
\node[load] (S) at (2.5,0.6) {\texttt{suffixes}$[2s_{16}{+}b_3]$};
\node[align=center, font=\scriptsize\itshape] (Lf) at (2.5,-0.05)
  {the suffix read waits\\for the branch read};
\draw[flow] (B) -- node[right]{$s_{16}$} (S);
% the rank table: three independent reads
\node (Rt) at (8.3,2.85) {\textbf{rank table}, 16~KiB};
\node[load] (R1) at (8.3,2.3) {\texttt{rows}$[2048r{+}i_1]$};
\node[load] (R2) at (8.3,1.45) {\texttt{rows}$[\mathrm{mix}(i_2)]$};
\node[load] (R3) at (8.3,0.6) {\texttt{rows}$[2048r_3{+}i_3]$};
\node[align=center, font=\scriptsize\itshape] (Rf) at (8.3,-0.05)
  {no read waits:\\issued together};
\begin{scope}[on background layer]
\node[mem, fit=(Lt)(P)(B)(S)(Lf), inner sep=4pt] (Lbox) {};
\node[mem, fit=(Rt)(R1)(R2)(R3)(Rf), inner sep=4pt] (Rbox) {};
\end{scope}
\node[box, minimum width=9.84cm] (y) at (5.4,-1.45)
  {24 values $y_j$, then $\times\,\mathrm{gain}[g] \times 1/\!\sqrt{16m}$};
\draw[flow] (w.south) -- ++(0,-0.2) -| (Lbox.north);
\draw[flow] (w.south) -- ++(0,-0.2) -| (Rbox.north);
\draw[flow] (Lbox.south) -- node[right, text height=1.6ex, text depth=0.4ex]{3 pattern bytes} (Lbox.south |- y.north);
\draw[flow] (Rbox.south) -- node[right, text height=1.6ex, text depth=0.4ex]{3 rank vectors} (Rbox.south |- y.north);
\end{tikzpicture}
\caption{The decode. The trellis reads return the Golay pattern bytes of the
three sections, and the row reads their rank vectors. Only the suffix read
waits, for the state $s_{16}$ that the branch read returns. The shell index
$m$ is recomputed from the decoded values, and $r_3$ comes from
Equation~\eqref{eq:xor}.}
\label{fig:decode}
\end{figure}

The entry of \texttt{inv\_norm} for $m=0$ is 0, so the origin, a legal word,
decodes to zeros with no special case. Trying every
codeword with both parities and every allowed row in each section gives
$m\le26$ and $\max|y_j|=10$.

Figure~\ref{fig:pseudo} gives the decode as code, and Table~\ref{tab:fields}
the fields of the word and the tables it reads.

\begin{figure}[t]
\small
\begin{verbatim}
decode(lo, hi16, rows, prefixes, branches, suffixes,
       inv_norm, gain, sigma_row):
  p  =  lo        & 1        # shared parity
  r  = (lo >>  1) & 1        # k-parity of section 1
  s8 = (lo >>  2) & 63       # trellis state at the first cut
  b1 = (lo >>  8) & 1        # prefix edge
  i1 = (lo >>  9) & 2047     # row index, section 1
  b2 = (lo >> 20) & 15       # branch edge
  i2 = ((lo >> 24) | ((hi16 & 7) << 8)) & 2047
  b3 = (hi16 >> 3) & 1       # suffix edge
  i3 = (hi16 >> 4) & 2047    # row index, section 3
  g  = (hi16 >> 15) & 1      # gain

  c1  = prefixes[2*s8 + b1]        # three pattern bytes
  br  = branches[16*s8 + b2]
  c2  = br & 255
  s16 = (br >> 8) & 63
  c3  = suffixes[2*s16 + b3]       # the one dependent read

  row1 = rows[2048*r + i1]         # three rank vectors,
  mid  = i2 < 1240                 #   issued together
  row2 = rows[mid ? i2 : 808 + i2]
  r3   = p ^ r ^ (mid ? 0 : 1)     # Equation (3), no stored bit
  row3 = rows[2048*r3 + i3]

  c, row = (c1, c2, c3), (row1, row2, row3)
  for k in range(24):              # y comes out in octad order
    s, j = k // 8, k % 8
    o    = p + 2 * bit(c[s], j)
    y[k] = val(o, rank(row[s], j))

  m = dot(y, y) / 16               # shell index, from y itself
  return y * gain[g] * inv_norm[m] * sigma_row
\end{verbatim}
\caption{Decoding one 48-bit word, given as its low 32 bits \texttt{lo} and
its high 16 bits \texttt{hi16}. \texttt{rank(row, j)} is bits $4j$ to $4j+3$
of the row, and \texttt{val(o, rank)} is the value at that rank in the list
of \S\ref{sec:state}. The 24 values come out octad by octad
(\S\ref{sec:octads}), so the dot product must pair \texttt{y[k]} with the
activation of the same coordinate. Pairing it with the activation at
\texttt{k} gives a wrong result and no error.}
\label{fig:pseudo}
\end{figure}

\begin{table}[t]
\centering
\small
\caption{The fields of the 48-bit word, and the tables the decode reads.}
\label{tab:fields}
\begin{tabular}{@{}llr l@{}}
\toprule
field & bits & width & what it selects \\
\midrule
$p$      & 0     & 1  & the shared parity \\
$r$      & 1     & 1  & the $k$-parity of section 1 \\
$s_8$    & 2--7  & 6  & the trellis state at the first cut \\
$b_1$    & 8     & 1  & the prefix edge \\
$i_1$    & 9--19 & 11 & the row of section 1 \\
$b_2$    & 20--23& 4  & the branch edge \\
$i_2$    & 24--34& 11 & the row of section 2 \\
$b_3$    & 35    & 1  & the suffix edge \\
$i_3$    & 36--46& 11 & the row of section 3 \\
$g$      & 47    & 1  & the gain level \\
\midrule
\multicolumn{4}{@{}l}{\textit{tables and constants}} \\
\texttt{rows}     & \multicolumn{2}{l}{$4{,}096\times$u32} & 16{,}384~B, 2{,}048 rows a class \\
$N_0$             & \multicolumn{2}{l}{1{,}240}            & class-0 rows of the middle list \\
\texttt{branches} & \multicolumn{2}{l}{$1{,}024\times$u16} & 2{,}048~B, byte $\mid$ next state \\
\texttt{prefixes} & \multicolumn{2}{l}{$128\times$u8}      & 128~B \\
\texttt{suffixes} & \multicolumn{2}{l}{$128\times$u8}      & 128~B \\
\texttt{inv\_norm}& \multicolumn{2}{l}{$32\times$f32}      & 128~B, $1/\sqrt{16m}$ \\
\midrule
total             & \multicolumn{2}{l}{}                   & 18{,}816~B \\
\bottomrule
\end{tabular}
\end{table}

\subsection{What the construction costs}
\label{sec:kept}

\Tetra{} and \Planes{} keep different sets of points of the same lattice.
Keeping the 2{,}048 cheapest rank vectors bounds each section on its own, so
the encoder searches a product of three 8-dimensional regions where \Planes{}
searched one 24-dimensional ball. This split makes the six-load decode
possible, and it costs quality.

Retention is the bit rate the theoretical limit would need to reach our error,
as a percentage of the rate we use. On 20{,}000 Gaussian blocks at 2 bits per
weight, in one process and with the same gain rule, \Tetra{} reaches
88.80\,\% retention and the \Planes{} codebook 91.98\,\%, so \Tetra{} has
9.2\,\% more mean squared error. On 20{,}000 rotated weight blocks of
Qwen3-0.6B's first layer, the gap is the same, 3.2 points. Sorting rank
vectors by $\mathrm{cost}(\rho)$ instead of exact squared distance accounts
for 0.6 points of it on Gaussian blocks.

Geometry predicts about as much. The shaping gain is how much the shape of the
searched region lowers the error against a cube. At best it is 0.7292~dB for a
product of three 8-dimensional balls and 1.0958~dB for one 24-dimensional
ball. The 0.3666~dB difference means 8.81\,\% more mean squared error for the
product. Our regions, cut by rank and by the trellis, are neither balls nor
continuous.

Every decoded point still lies in $\Lambda_{24}$. Its pattern bytes follow a
trellis path, so they form a Golay codeword, and Equation~\eqref{eq:xor} gives
the $\sum k$ parity. Three separate quantizers
on $E_8$, the best lattice quantizer known in 8 dimensions, would keep neither
constraint. At the same density of points, the $E_8$ lattice has 9.0\,\% more mean
squared error than $\Lambda_{24}$: $G(E_8)/G(\Lambda_{24}) = 0.071682/0.065771
= 1.0899$, where $G$ is the normalized second
moment~\citep[Table~2.3]{conway1999sphere}.

Our data cannot separate the two codebooks on a model. On Qwen3-4B, with the
same evaluation code, bare \Tetra{} (without the changes of \S\ref{sec:model})
scores 2.11 MMLU points below \Planes{} on a 2{,}280-question sample,
$[-0.95, +5.17]$, $p=0.11$, with 4.64\,\% lower perplexity. Each file comes
from one calibration draw, and three draws of the \Planes{} encoding spread
MMLU over 5.83 points on this sample, a standard deviation of 2.92. A month of
encoder changes also separates the two files. Re-encoding the first layer of
the \Planes{} file with the newer code changes 87\,\% of its indices.

\section{From the codebook to a served file}
\label{sec:model}

A served file holds more than lattice codes. The three changes below start
from an encoding that already stores \texttt{v\_proj} in int4
(\S\ref{sec:int4}), and none of them changes the decoder or a remaining lattice
code.

\subsection{What a file holds, and how we count it}
\label{sec:file}

Each layer of the dense Qwen3 models~\citep{qwen3_2025} has seven projection
matrices: $q$, $k$, $v$ and $o$ in attention, and gate, up and down in the
feed-forward block. The 4B and 8B models have 36 layers (252 projections), and
the 14B has 40 (280).

Each projection is stored as \Tetra{} blocks or as an int4 record: 4-bit
weights in groups of 128, with an f16 scale and offset per group, read by a
separate GPU kernel. A \Tetra{} projection also stores one scale per output
row. It keeps unquantized the tail of each row, the weights that do not fill a
whole block, in f32 in the file and f16 on the GPU. The RMSNorm weights stay in
f16. The embedding table is stored separately, and so is the output head at 8B
and 14B. At 4B the head is tied to the embedding.

We count memory in bits per parameter over the whole model. That is all the
weight bits the served model keeps, embedding, tail, scales and int4 records included,
over the number of parameters. For our files, the \texttt{rtbits} tool
computes it from the file's records, at the widths the GPU holds. For AWQ and
IQ2\_XXS, it is the checkpoint's size in bits over the number of parameters.
The KV cache and the activations are not counted.

\subsection{Training the row scales}
\label{sec:rowscales}

The encoder sets each row scale $\sigma_{\mathrm{row}}$ of
Equation~\eqref{eq:recon} from the norm of the rotated row. We retrain only
these scales, with the FP16 model as teacher~\citep{hinton2015} and a KL loss
on the output probabilities over DCLM-Edu text~\citep{dclmedu}. The trained
scales replace the old ones in place, so no bit is added. At 4B the training
sees 19.5 million tokens and takes 1.7~h on one L40S. The 8B and 14B runs take
1.1~h and 1.8~h on one H200.

\subsection{int4 where the lattice loses the most}
\label{sec:int4}

Every file of Table~\ref{tab:chain} stores \texttt{v\_proj} in int4. Qwen3
uses grouped-query attention~\citep{gqa2023}, so these matrices are small. At
4B, in an earlier encoding calibrated on C4 text, moving them from \Tetra{} to
int4 gained 1.73 MMLU points $[0.87, 2.59]$ on the full test set, for 0.05
bits per parameter.

The three files we serve, which we call \emph{sealed}, also store
\texttt{down\_proj} in int4 in a range of middle layers, and the 4B and 14B
files \texttt{o\_proj} as well (Table~\ref{tab:files}). \S\ref{sec:chain}
explains why the 8B does not.

\subsection{4-bit embedding tables}
\label{sec:q4emb}

Qwen3's embedding table has 151{,}936 rows and is 9.7\,\% of the parameters at
4B. At 8B and 14B the output head is a second table of the same shape, and the
two hold 15.2\,\% and 10.5\,\% of the parameters. Before this step our files
stored them in f16, and the served model quantized them to 8 bits at load. The
sealed files store them in 4 bits like the int4 projections, but in groups of
64 as MLX does~\citep{mlx2023}. The GPU dequantizes rows on the
fly, for the input tokens and inside the head's matrix-vector product. Going
from 8 to 4 bits frees 0.19, 0.62 and 0.78~GB at 4B, 8B and 14B. This saving
pays for the int4 projections, and every sealed file needs fewer bits per parameter
than the file it was built from (\S\ref{sec:chain}).

\begin{table}[t]
\centering\small
\caption{The three sealed files. Layers are numbered from 0: ``down 12--23''
means \texttt{down\_proj} in layers 12 to 23. Bits per parameter are counted as
in \S\ref{sec:file}. The file is larger because it also carries the tokenizer and the
config (11.4~MB), and stores the tail in f32 and the row scales in f64. Data:
paper2-chain.csv; the sealing logs (\texttt{rtbits} and \texttt{int4swap}
outputs) for the splits, the windows and the file sizes.}
\label{tab:files}
\begin{tabular}{lrrr}
\toprule
 & Qwen3-4B & Qwen3-8B & Qwen3-14B\\
\midrule
projections, \Tetra{} $+$ int4 & 168 $+$ 84 & 199 $+$ 53 & 181 $+$ 99\\
int4 projections & $v$, $o$, down 12--23 & $v$, down 10--26 & $v$, $o$, down 10--28\\
4-bit embedding tables & 1 (tied head) & 2 & 2\\
file size (bytes) & 1{,}418{,}224{,}685 & 2{,}815{,}098{,}745 & 5{,}087{,}000{,}541\\
bits per parameter, whole model & 2.7320 & 2.6953 & 2.7305\\
\bottomrule
\end{tabular}
\end{table}

\section{The kernel}
\label{sec:kernel}

\subsection{Decoding inside the matrix-vector product}

The kernel keeps the thread layout of our earlier one~\citep{llvq1preprint}.
One warp (threads that run in lockstep) computes each output row, and a
256-thread block holds 8 rows. Each thread of the warp, or lane, decodes one
block of weights at a time. The activation goes into shared memory in tiles of
$T$ blocks ($96T$ bytes), with two barriers per tile. A shuffle reduction sums
the lanes' results, and a last step multiplies by $\sigma_{\mathrm{row}}$ and
adds the unquantized tail. Only the block decode differs.

Decoding a block takes six table loads and two small lookups, for the gain and
the norm. Three choices make it fast. First, each coordinate value is held in a
register table as a biased byte, $\mathrm{val}+128$. Placed in the low byte of
\texttt{0x4b0000\_\_}, it reads exactly as the float $2^{23}+\mathrm{byte}$,
and one subtraction gives the value back. A coordinate thus costs one byte
permute, one add and one fused multiply-add, with no integer-to-float
conversion. Second, every loop is unrolled with constant indices, and the
compiler reports 40 registers and no spill to local memory for the benchmark kernel, as
for \Planes{}. Third, no branch depends on the data, so lanes never diverge.

Counted from the source, a block's rank decode and 24 multiply-adds take about
150 instructions (about 60 on the FMA pipe, about 85 on the ALU pipe and six
loads), and the gain and the norm about 20 more. A decoder that extracts and
converts each coordinate needs about 380, plus 24 conversions.

The kernel reads the 48 bits the file stores for each block, and no code is
expanded. Loading only reorders the bits of each word and pads each row to a
multiple of 8~bytes. In the same process, the sealed 4B file loads in 4.1~s
and its dense reconstruction (the file decoded to f16) in 72.9~s.

\paragraph{The rest of a served model.} The int4 projections use a separate
kernel with the same thread layout. It dequantizes each weight with its group's
scale and offset and accumulates in f32. It stages the whole activation in
shared memory. A 14B \texttt{down\_proj} needs 69{,}632~B, more than the
48~KiB default, so the host opts in to that amount, under the L40S limit of
101{,}376~B~\citep{cudaguide}. Two small kernels serve the 4-bit embedding
tables, and a third rotates the activations online, 96 times a token at 4B,
once per group of projections that share a rotation.

\paragraph{How the kernel is checked.} Before any timing, we recompute in f64,
from the same decoded weights, every output row of the 4B shapes in every
benchmark arm that reads weights: 1{,}105{,}920 rows per arm.
Each must match the kernel's row within $10^{-5}\cdot\sum|w\cdot x|$, or
$10^{-3}$ for AWQ and cuBLAS, which write f16. The worst \Tetra{} row is off by
$2.0\times10^{-8}\cdot\sum|w\cdot x|$. The decoder's CUDA source, compiled for
the CPU, gives the same 24 lattice coordinates as an independent Rust decoder,
bit for bit, on 4{,}000 test words: the origin and the all-ones word, each
with and without the gain bit, and random words.

End to end, the kernel and the dense reconstruction agree until two logits
nearly tie. On one prompt and in one process, they generate the same 256
greedy tokens at 4B and 8B for each sealed file. At both sizes the text falls into a
one-sentence loop. At 14B they first differ at token 78, and at token 137 when
both run with f16 embedding tables. A second check runs 203 prompt tokens
through the kernel, in one call and in 203 single-token calls. Both give the
same next token at all three sizes. The two paths still move a logit by up to
0.095, 1.35 and 0.79 at 4B, 8B and 14B, for logits up to 30.4, 36.6 and 27.5.
The divergence at token 78 is consistent with a near tie between two logits,
flipped by differences of this size, but we did not measure the margin there.
Our preregistration counts a first divergence at or after token 32 as not a
defect.

\subsection{Bytes and time on the projections}
\label{sec:bench}

\begin{table}[t]
\centering\small\setlength{\tabcolsep}{5pt}
\caption{Kernel benchmark on the 4B shapes at tile $T=64$: one process, one
L40S, seven rounds with the first two dropped. Every arm times the same 252
matrices, and the \Tetra{} arm reads a 4B file with no int4 projection. ``No
weights'' has the same thread layout and reads no block code, only the tail
and the row scales. The lattice arms hold the tail in f32, as the files store
it, where the served kernel uses f16. The \Tetra{} row leaves out 4.2~MB of
row padding, 0.009~\bw{}. The AWQ arm is its authors' GEMV
kernel~\citep{awq2024} ported to our harness, not the Marlin kernel that vLLM
serves. Its row includes 10.1~MB of padding in that kernel's scale and zero
buffers, 0.022~\bw{}. Without it, w4g128 costs 4.156~\bw{}. Ten arms:
Appendix~\ref{app:bench}. Data: echelle-formats.csv.}
\label{tab:bench}
\begin{tabular}{lrrrrr}
\toprule
 & FP16 & AWQ w4g128 & \Planes{} & \Tetra{} & no weights\\
\midrule
median ms & 11.022 & 3.263 & 5.004 & 3.734 & 2.294\\
GB read per pass & 7.27 & 1.90 & 2.18 & 0.98 & 0.07\\
\bw{} in VRAM & 16.000 & 4.179 & 4.804 & \textbf{2.150} & 0.159\\
GB/s, fastest round & 660 & 582 & 436 & 262 & 32\\
\bottomrule
\end{tabular}
\end{table}

Per weight, \Tetra{} reads about half the bits of 4-bit AWQ and of \Planes{},
yet its time is still above AWQ's (Table~\ref{tab:bench}): reading half the
bytes does not halve the time.

\paragraph{The activation tile.}\label{sec:tile} The size of the activation
tile matters to \Tetra{} and hardly to the other arms. The tile $T$ changes
no output bit and costs no storage. On an Ada GPU (sm\_89), it sets how much of each
SM's 128~KiB pool of L1 and shared memory is left to cache the decoder
tables~\citep{cudaguide}. Occupancy does not move. Six 256-thread blocks fit at
every tile, filling the 1{,}536 threads an SM holds, and shared memory never
binds. At $T = 128$, 64 and 32, one process per tile on one L40S, the median
pass takes 2.281, 2.291 and 2.392~ms with no weights, 5.068, 5.011 and
5.047~ms for \Planes{}, and 4.353, 3.701 and 3.841~ms for \Tetra{} (data:
tuile-l40s.csv). Each arm's range, largest over smallest, is 4.9, 1.1 and
17.6\,\%. The three arms share the thread layout, the activation copy, the
barriers and the occupancy. \Planes{} reads one entry per block from a 12~KiB
table, \Tetra{} seven from 18.4~KiB. We attribute \Tetra{}'s larger range to
pressure on L1 but have not proven it, because our rented platform exposes no
performance counters. We serve $T=64$ on sm\_89. On an sm\_120 card, an
earlier sweep over the same 252 matrices puts \Tetra{} fastest at $T=32$ and
\Planes{} at $T=128$, with \Tetra{} at 0.82 to 1.03 times the speed of
\Planes{} depending on the tile. On an A100 (\S\ref{sec:a100}), \Tetra{} is
fastest at $T=128$.

\subsection{One launch per token}
\label{sec:4e}

\begin{figure}[t]
\centering
\begin{tikzpicture}[node distance=3pt and 2.5pt]
\node[title, anchor=west] (ta) at (0,0)
  {Per-matrix path: one launch for each operation, back to the host each time};
\node[box, anchor=north west] (a1) at ([yshift=-3pt]ta.south west) {norm};
\node[box, right=of a1] (a2) {rotate};
\node[box, right=of a2] (a3) {$q$};
\node[box, right=of a3] (a4) {$k$};
\node[box, right=of a4] (a5) {$v$};
\node[box, right=of a5] (a6) {attention};
\node[box, right=of a6] (a7) {$o$};
\node[box, right=of a7] (a8) {norm};
\node[box, right=of a8] (a9) {rotate};
\node[box, right=of a9] (a10) {gate};
\node[box, right=of a10] (a11) {up};
\node[box, right=of a11] (a12) {down};
\node[note, anchor=west] at ([xshift=6pt]a12.east) {$\times$ 36 layers, then the head};

\node[title, anchor=north west] (tb) at ([yshift=-11pt]a1.south west)
  {Persistent kernel: one launch for the whole token};
\node[mem, anchor=north west] (b1) at ([yshift=-7pt,xshift=4pt]tb.south west) {norm,\\rotate};
\node[mem, right=8pt of b1] (b2) {$q$, $k$, $v$\\rows};
\node[mem, right=8pt of b2] (b3) {RoPE,\\KV append};
\node[mem, right=8pt of b3] (b4) {attention};
\node[mem, right=8pt of b4] (b5) {$o$\\rows};
\node[mem, right=8pt of b5] (b6) {norm,\\rotate};
\node[mem, right=8pt of b6] (b7) {gate, up\\rows};
\node[mem, right=8pt of b7] (b8) {down\\rows};
\foreach \a/\b in {b1/b2,b2/b3,b3/b4,b4/b5,b5/b6,b6/b7,b7/b8}
  \draw[flow, okBlue] (\a) -- (\b);
\begin{scope}[on background layer]
\node[panel, draw=okBlue, fit=(b1)(b8)] (pb) {};
\end{scope}
\node[note, anchor=west] at ([xshift=5pt]pb.east) {$\times$ 36 layers,\\then the head\\and its argmax};
\node[note, anchor=north west, align=left] at ([yshift=-3pt]pb.south west)
  {Each arrow is a grid-wide barrier. 142 blocks of 512 threads, one per SM, claim rows from a shared counter.};
\end{tikzpicture}
\caption{One token of the 4B in the two engines. The per-matrix path returns
to the host between operations. The persistent kernel stays on the card for
the whole token, 218 barriers at 4B, and computes each row with the same
products in the same order.}
\label{fig:4e}
\end{figure}

The served engine runs a whole token, every layer and the output head, in one
cooperative launch (Figure~\ref{fig:4e}). The kernel above is launched once
per matrix, so it leaves the card between any two operations of a token: 252
projections at 4B, plus the norms, the rotations and the attention. Our engine
first served the files that way.

The grid has one 512-thread block per SM, 142 on the L40S. A token is a fixed
sequence of phases, and between two phases every block waits at a grid-wide
barrier, 218 times a token at 4B. Inside a phase the blocks claim output rows
from a shared counter. A \Tetra{} row is computed as in the per-matrix
kernel: the same lanes, the same block order, the same 24
products in the same order and the same last step. In the three places where
a compiler could fuse a multiply with an add, the operations are written out
by hand. The int4 rows read the same words, laid out once at load so that the
four words a lane needs from a tile sit side by side and one 128-bit load
fetches them. The card holds those bytes once.

\paragraph{Same tokens.} The persistent kernel must return the tokens of the
per-matrix path, and each load checks it on the card before serving. In the
runs behind our speeds it returned the same 256 greedy tokens at 4B, 8B and
14B, after prompts of 5, 2{,}048, 4{,}096 and 8{,}192 tokens.

\paragraph{Attention.}\label{sec:attn} Qwen3 shares each key and value head
among several query heads. The \emph{expanded} form copies each cached head
to its query heads before the product, and every score in this paper ran
under it. The served engine reads each cached head once, in an attention
kernel of its own. Its sums run in another order, so its logits are not bit
for bit those of the expanded form. We serve it under a tolerance fixed in
advance: paired WikiText-2 perplexity within $\pm0.12\,\%$ of the reference
form, which is what this paired test resolves on one file. Reading each head
once moves perplexity by $+0.0095$, $-0.0048$ and $+0.0023\,\%$ at 4B, 8B and
14B. The attention kernel moves it by a further $+0.0018$, $+0.0043$ and
$-0.0108\,\%$, over 12 windows at 4B and 6 at 8B and 14B. Every interval lies
inside the band.

\paragraph{Prompts.} The per-matrix path prefills a prompt in chunks of
4{,}096 positions. At 4B the served engine instead feeds the prompt through a
multi-row pass of the persistent kernel, the \emph{chain prefill}. It reads
the served words in place, and its paired perplexity is $+0.0007\,\%$ from the
per-matrix prefill's. That pass does not yet carry the 8B and the 14B
(\S\ref{sec:context}).

\paragraph{Limits.} The persistent kernel decodes one request at a time, with
greedy sampling, on the three sealed shapes. It exists for NVIDIA cards only
and is measured on the L40S only.

\needspace{8\baselineskip}
\section{Experiments}
\label{sec:experiments}

\subsection{Setup}
\label{sec:setup}

\paragraph{Models and encoding.} Qwen3-4B, 8B and 14B~\citep{qwen3_2025} are
each encoded with the first 131{,}072 tokens of DCLM-Edu~\citep{dclmedu}, the
same text at every size, then trained and assembled as in \S\ref{sec:model}.

\paragraph{Quality.} The benchmark is MMLU~\citep{mmlu2021}, 5-shot,
micro-averaged over all 14{,}042 questions of its test split, with the answer
read from the logits of the four answer letters. Each file is scored on its
dense reconstruction in an ordinary forward pass. The kernel computes with the
same decoded weights, and the checks of \S\ref{sec:kernel} tie the two. At 4B,
we scored the file one step before sealing, with the same int4 matrices
rebuilt at load by the same quantizer. On 57 questions, the sealed 4B file
gives the same answers and logits. Our gaps to FP16 and AWQ, and between our
own files, are paired on the same questions, with a 95\,\% interval from a
bootstrap that resamples questions within each subject (10{,}000 draws) and an
exact McNemar test. On the full test set these intervals are 0.7 to 1.7 points
wide, so how small a gap this test separates depends on the pair. A gap whose interval contains zero is \emph{not
separated}. We ran no equivalence test, so this does not mean the two scores
are equal.

\paragraph{Perplexity.} We also read each file's perplexity on the raw test
split of WikiText-2~\citep{wikitext2016}, over 12 non-overlapping windows of
4{,}096 tokens, on the dense reconstruction in f16, as for MMLU, on the L40S.
Each value is one run. Two files read on the same windows give per-window
losses that pair, and we give such pairs a 95\,\% $t$ interval over the 12
windows.

\paragraph{Reasoning.} To score generated reasoning, not one logit per
question, we add all 1{,}319 test problems of GSM8K~\citep{gsm8k2021},
zero-shot in the model's chat template with its thinking block left empty
(Qwen3's non-thinking mode~\citep{qwen3_2025}). The prompt asks for the answer
in a \texttt{\textbackslash{}boxed\{\}}. Decoding is greedy and stops at the
end of the turn or after 1{,}024 tokens. The answer is the last number in the
last box, or the last number of the text when no box closes. Unlike MMLU, our
files are scored through the kernel, on the per-matrix path. FP16 and AWQ generate in vLLM from
the same prompt tokens, and one grader scores every arm. Gaps are paired, with
a 95\,\% normal interval and an exact McNemar test.

\paragraph{Scores and the served engine.} Every score in this paper was
measured with the expanded attention (\S\ref{sec:4e}). The served engine
differs from that form only by its attention, which a perplexity tolerance
ties to the expanded one, and by the persistent kernel, which returns the per-matrix
tokens. As a direct check, 50 GSM8K problems per size ran through the
persistent kernel: 40, 46 and 46 right at 4B, 8B and 14B, the counts of the
earlier engine on those problems.

\paragraph{Speed and memory.} Decode speed is measured on one NVIDIA L40S at
batch 1 with greedy sampling. Our files run in our engine under their served
configuration (\S\ref{sec:4e}) and generate 256 tokens after a 5-token prompt.
Their speed is tokens over wall-clock time, prompt included. FP16 and AWQ run
in vLLM 0.26.0~\citep{vllm2023}, AWQ with the Marlin kernel, and generate 128
tokens, timed the same way. Both engines report the median of five timed
rounds, after one warm-up round in ours and two in vLLM, and the five rounds
of every median lie within 2\,\% of each other. IQ2\_S runs in
llama.cpp~\citep{llamacpp} at commit \texttt{cd959bae9}, built for the card,
with \texttt{llama-bench}: 256 tokens after a prompt of 5 and of 8{,}192
tokens, as the median of five rounds. Its speed leaves the prompt out.
IQ2\_XXS at 4B keeps its earlier measurement, with build b10689 on 128 tokens
from an empty prompt, as the mean of five repetitions. The memory of
Table~\ref{tab:main} counts only weight bytes (\S\ref{sec:file}): GPU buffers
for our files, weight files for the others. Card memory
(\S\ref{sec:duel}, \S\ref{sec:context}) is read from the driver every 100~ms,
as the peak minus the idle level.

\paragraph{Baselines.} FP16 is Qwen's published checkpoint, stored in bf16 and
run in f16. AWQ is Qwen's official w4g128 checkpoint at each size, scored on
MMLU in our harness after conversion to f16 and timed in vLLM on the packed
checkpoint. IQ2\_S is the llama.cpp format nearest our bits per parameter. A
preregistered rule picked it from the file sizes, before any score. We built
it with \texttt{llama-quantize} and an importance matrix from the DCLM-Edu
text of our files. Its perplexity is read on our 12 windows. As a gate, FP16
through llama.cpp reads within 0.05\,\% of FP16 in our engine at every size.
Its MMLU and GSM8K go through \texttt{llama-server} with our prompts and our
grader. IQ2\_XXS at 4B is an earlier build, with an importance matrix from
131{,}072 tokens of C4, scored the same way. We cite the
MMLU of QTIP, QuIP\# and the original LLVQ from \citet{llvq2026} and do not
run them.

\subsection{Main results}
\label{sec:main}

\begin{table}[t]
\centering\small\setlength{\tabcolsep}{3.2pt}
\caption{Quality, speed and memory at three sizes, on one L40S, with the
protocols of \S\ref{sec:setup}; ppl is the WikiText-2 perplexity. The engine
named in each row times the speed. For our files, weights GB are the buffers
on the card, row padding included. $^{a}$The sealed file decoded to f16 and run as an ordinary model,
in the same process as the kernel. Its speed is that of any f16 model of this
size on the per-matrix path, which copies the output head at
every token. This row has no MMLU of its own. $^{b}$Fine-tuned rows of Table~6 of \citet{llvq2026}, from
their harness and MMLU protocol. $^{c}$Bits per quantized weight, as the source
gives them. It gives no rate for the whole model. $^{d}$Bits per parameter
from the file's bytes, speed of the decode alone after a 5-token prompt, GB of
the file. Table~\ref{tab:duel} puts our files in the same accounting. Data:
paper2-results.csv, paper2-ppl.csv.}
\label{tab:main}
\begin{tabular}{llrrrrr}
\toprule
 & format and engine & b/param & MMLU & ppl & decode tok/s & weights GB\\
\midrule
\multirow{9}{*}{4B}
 & FP16, vLLM & 16.00 & 70.14 & 12.24 & 83.1 & 8.04\\
 & f16 dense path, our engine$^{a}$ & 16.00 & & & 43.0 & 8.04\\
 & AWQ w4g128, vLLM & 5.30 & 68.14 & 13.52 & 200.5 & 2.67\\
 & IQ2\_XXS, llama.cpp & 2.48 & 39.78 & & 312.9 & 1.25\\
 & IQ2\_S, llama.cpp$^{d}$ & 2.82 & 54.29 & 16.78 & 287.5 & 1.42\\
 & \textbf{\Tetra{}, our engine} & \textbf{2.73} & \textbf{63.37} & \textbf{12.58} & \textbf{221.5} & \textbf{1.38}\\
 & \textcolor{okGray}{LLVQ, cited$^{b}$} & \textcolor{okGray}{2 b/w$^{c}$} & \textcolor{okGray}{62.8} & & & \\
 & \textcolor{okGray}{QTIP 3INST, cited$^{b}$} & \textcolor{okGray}{2 b/w$^{c}$} & \textcolor{okGray}{59.5} & & & \\
 & \textcolor{okGray}{QuIP\#, cited$^{b}$} & \textcolor{okGray}{2 b/w$^{c}$} & \textcolor{okGray}{52.9} & & & \\
\midrule
\multirow{5}{*}{8B}
 & FP16, vLLM & 16.00 & 75.05 & 8.99 & 46.3 & 16.38\\
 & f16 dense path, our engine$^{a}$ & 16.00 & & & 26.4 & 16.38\\
 & AWQ w4g128, vLLM & 5.96 & 73.79 & 9.42 & 123.2 & 6.10\\
 & IQ2\_S, llama.cpp$^{d}$ & 2.80 & 65.45 & 10.75 & 208.4 & 2.86\\
 & \textbf{\Tetra{}, our engine} & \textbf{2.70} & \textbf{69.58} & \textbf{9.71} & \textbf{142.5} & \textbf{2.76}\\
\midrule
\multirow{5}{*}{14B}
 & FP16, vLLM & 16.00 & 78.88 & 7.98 & 25.8 & 29.54\\
 & f16 dense path, our engine$^{a}$ & 16.00 & & & 16.9 & 29.54\\
 & AWQ w4g128, vLLM & 5.40 & 78.12 & 8.29 & 77.8 & 9.98\\
 & IQ2\_S, llama.cpp$^{d}$ & 2.69 & 72.77 & 9.47 & 123.5 & 4.96\\
 & \textbf{\Tetra{}, our engine} & \textbf{2.73} & \textbf{75.66} & \textbf{8.44} & \textbf{86.3} & \textbf{5.04}\\
\bottomrule
\end{tabular}
\end{table}

\begin{table}[t]
\centering\small\setlength{\tabcolsep}{6pt}
\caption{Paired gaps of our files to the baselines. MMLU: the baseline minus
our file, in points on the 14{,}042 questions, with 95\,\% intervals.
Perplexity: our file relative to the baseline over the 12 windows, where a
negative value means our file reads lower. Data: paper2-gaps.csv,
paper2-ppl.csv.}
\label{tab:gaps}
\begin{tabular}{lllrr}
\toprule
 & \multicolumn{2}{c}{MMLU, points below} & \multicolumn{2}{c}{perplexity against}\\
\cmidrule(lr){2-3}\cmidrule(lr){4-5}
 & FP16 & AWQ & FP16 & AWQ\\
\midrule
4B & 6.77 $[6.05, 7.50]$ & 4.76 $[4.02, 5.49]$ & $+2.8\,\%$ & $-6.9\,\%$\\
8B & 5.48 $[4.83, 6.10]$ & 4.21 $[3.54, 4.86]$ & $+8.0\,\%$ & $+3.0\,\%$\\
14B & 3.22 $[2.69, 3.75]$ & 2.46 $[1.92, 3.00]$ & $+5.8\,\%$ & $+1.9\,\%$\\
\bottomrule
\end{tabular}
\end{table}

\begin{figure}[t]
\centering
\includegraphics{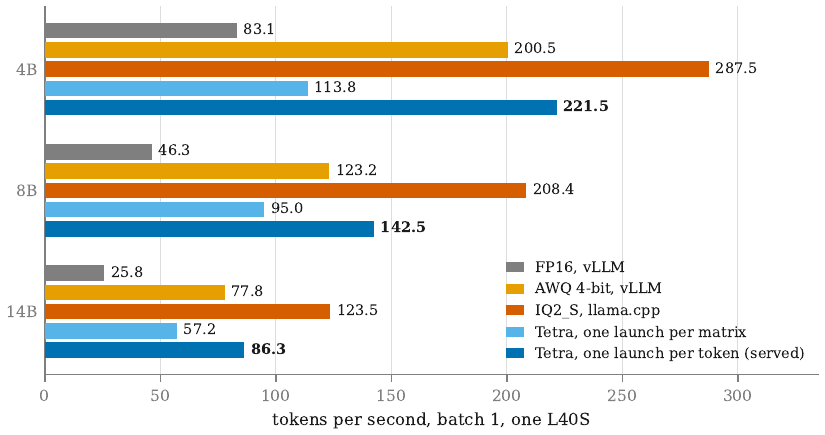}
\caption{Decode speed at batch 1 on one L40S, from Table~\ref{tab:main}. The
three engines differ, so the bars compare what a user gets, not the formats.
The IQ2\_S bar is the decode alone. Our decode alone reads 227.4, 144.5 and
90.0 (Table~\ref{tab:duel}). Data: paper2-results.csv.}
\label{fig:speed}
\end{figure}

\paragraph{Quality.} Our files lose quality to both baselines, and the loss
shrinks as the model grows (Table~\ref{tab:gaps}). On the same questions, the
gap to FP16 falls from 6.77 MMLU points at 4B to 3.22 at 14B, and the gap to
AWQ from 4.76 to 2.46. The trend is not established at every step: the
0.56-point fall of the gap to AWQ from 4B to 8B is not separated from zero
($z=1.1$). AWQ itself sits 2.00 $[1.50, 2.51]$, 1.27 $[0.83, 1.70]$ and 0.75
$[0.40, 1.12]$ points below FP16.

\paragraph{Perplexity.} Perplexity ranks the files differently from MMLU. It
puts our 4B 6.9\,\% below AWQ, where MMLU puts it behind, and it finds the gap
to AWQ smallest at 4B, where MMLU finds it smallest at 14B
(Table~\ref{tab:gaps}). Against FP16 our files read 2.8 to 8.0\,\% higher. All
six gaps are separated from zero, paired over the 12 windows. AWQ's own
perplexity is 10.5\,\% above FP16's at 4B.

\paragraph{Memory.} Our files need about half of AWQ's bits per parameter, 45
to 52\,\%, and part of that comes from the embedding tables. Ours are in 4
bits, while AWQ keeps its embedding and head in f16. These tables weigh most
at 8B (\S\ref{sec:q4emb}), which is where AWQ needs the most bits per
parameter. If AWQ stored them in 4 bits like ours, our files would need 64 to
65\,\% of its bits per parameter. We computed this from the bytes and did not
measure AWQ's quality with such tables.

\paragraph{Against other 2-bit formats.} IQ2\_S, the llama.cpp format nearest
our bits, scores below our files on MMLU and perplexity at every size and
decodes faster; \S\ref{sec:duel} gives that comparison. At 4B, IQ2\_XXS scores
23.6 points below our file for 0.25 fewer bits per parameter. The cited rows
of Table~\ref{tab:main} come from another harness and count bits per quantized
weight, so they place our 63.37 only roughly and do not rank it. It is close
to the 62.8 of the original LLVQ with fine-tuning, and above the 59.5 of QTIP
with its 3INST code.

\paragraph{Speed.} One launch per token makes the files 1.4 to 1.7 times
faster than the per-matrix path, for the same tokens
(Figure~\ref{fig:speed}). The sealed files decode 221.5, 142.5 and 86.3 tokens
per second, against 129.0, 103.0 and 62.8 for the per-matrix path under the
same configuration. Formed round by round in one job, the ratio is 1.717
$[1.714, 1.722]$ at 4B, 1.384 $[1.382, 1.385]$ at 8B and 1.375
$[1.369, 1.376]$ at 14B. With the expanded attention, our first
configuration, the per-matrix path reads 113.8, 95.0 and 57.2
(\S\ref{sec:4e}). After loading, the card holds 1.98, 3.36 and 5.64~GB, which
is the weights plus the CUDA context and the persistent kernel's buffers.

The 4-bit output head explains a large part of the files' lead over their own
dense reconstruction. In the first configuration, with the embedding and head
in f16 on both sides, the files decoded 55.0, 38.9 and 27.3 tokens per second
against 43.3, 26.4 and 16.9 for the dense arm of the same runs: 1.27, 1.47 and
1.61 times, from unrounded medians.

We form no ratio between our speeds and those of vLLM or llama.cpp, because
such a ratio would mix the format with the engine. vLLM, for one, runs FP16
faster than our dense path does. In levels, our files now decode above AWQ in
vLLM at every size, and below IQ2\_S in llama.cpp.

\subsection{Reasoning: GSM8K through the kernel}
\label{sec:gsm8k}

\begin{table}[t]
\centering\small\setlength{\tabcolsep}{5pt}
\caption{GSM8K, with the protocol of \S\ref{sec:setup}. The gaps are paired,
with 95\,\% intervals. Data: paper2-gsm8k.csv, paper2-gsm8k-gaps.csv.}
\label{tab:gsm8k}
\begin{tabular}{lrrrll}
\toprule
 & FP16 & AWQ & \Tetra{} & FP16 minus \Tetra{} & AWQ minus \Tetra{}\\
\midrule
4B & 92.12 & 89.01 & \textbf{82.49} & 9.63 $[7.69, 11.57]$ & 6.52 $[4.39, 8.65]$\\
8B & 93.25 & 92.95 & \textbf{88.63} & 4.62 $[3.00, 6.25]$ & 4.32 $[2.83, 5.81]$\\
14B & 95.30 & 95.38 & \textbf{92.04} & 3.26 $[1.93, 4.59]$ & 3.34 $[2.12, 4.55]$\\
\bottomrule
\end{tabular}
\end{table}

On generated reasoning, our files lose to both baselines at every size. All
six gaps are separated from zero, with every $p$ below $10^{-5}$
(Table~\ref{tab:gsm8k}). AWQ is harder to tell from FP16 on this test: it sits
3.11 points below at 4B, then 0.30 and $-0.08$ at 8B and 14B, which the test
set cannot separate from zero.

Reasoning costs our files more than multiple choice does, although the two
ways of counting disagree on where. In points, the loss is larger on GSM8K
only at 4B, 9.63 against 6.77, where the GSM8K interval excludes the MMLU gap.
At 8B and 14B the intervals contain the MMLU gaps of 5.48 and 3.22. In errors,
the sealed files make 2.22, 1.69 and 1.69 times as many as FP16 on GSM8K,
against 1.23, 1.22 and 1.15 times on MMLU, so the cost is higher at every
size.

\paragraph{Two engines.} The engine does not explain these gaps. We ran the
FP16 4B checkpoint through our dense path, and it scored 91.51 against 92.12
in vLLM, $-0.61$ points $[-1.27, 0.06]$, $p=0.12$. Against this same-engine
FP16, the 4B file loses 9.02 points $[7.07, 10.97]$, against 9.63 across
engines. On the 50 problems of a pilot, the kernel on the L40S and the dense
reconstruction on an Apple GPU give the same 50 answers for the 4B file.

\subsection{At the same bits: llama.cpp IQ2\_S}
\label{sec:duel}

\begin{figure}[t]
\centering
\includegraphics{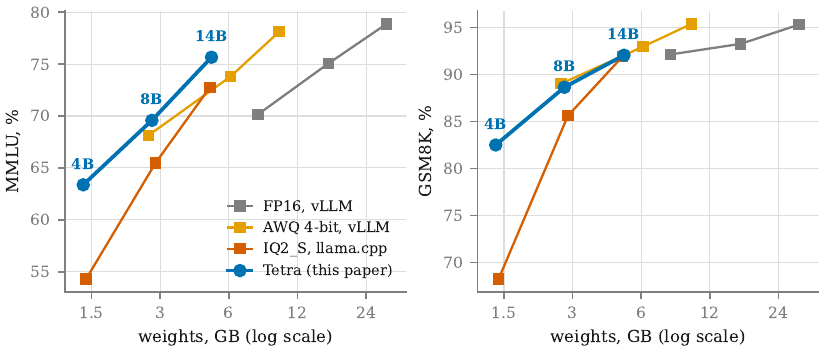}
\caption{Quality against weight memory. Each line joins the 4B, 8B and 14B of
one format; up and to the left is better. Left: MMLU. Right: GSM8K. Weights as
in Table~\ref{tab:main}. IQ2\_XXS at 4B (39.78 MMLU) is below the axis. Data:
paper2-results.csv, paper2-gsm8k.csv.}
\label{fig:frontier}
\end{figure}

\begin{table}[t]
\centering\small\setlength{\tabcolsep}{3.5pt}
\caption{Our files against llama.cpp on one L40S, one day, one accounting:
bits per parameter from file bytes, speed of the decode alone as the median of
five rounds, GB/s as file bytes times tokens per second, card memory as the
peak after a 5-token prompt. IQ2\_S is ours, built as in \S\ref{sec:setup};
UD-IQ2\_XXS is a community file one rung below. Our quality columns repeat
Tables~\ref{tab:main} and~\ref{tab:gsm8k}. Data: paper2-duel.csv.}
\label{tab:duel}
\begin{tabular}{llrrrrrrrr}
\toprule
 & & & & & & \multicolumn{2}{c}{decode tok/s after} & GB/s & card GB\\
 & format & b/param & ppl & MMLU & GSM8K & 5 & 8{,}192 & after 5 & after 5\\
\midrule
\multirow{3}{*}{4B}
 & \textbf{\Tetra{}} & \textbf{2.82} & \textbf{12.58} & \textbf{63.37} & \textbf{82.49} & \textbf{227.4} & \textbf{160.8} & \textbf{322} & \textbf{2.01}\\
 & IQ2\_S & 2.82 & 16.78 & 54.29 & 68.23 & 287.5 & 195.4 & 408 & 2.17\\
 & \textcolor{okGray}{UD-IQ2\_XXS} & \textcolor{okGray}{2.50} & \textcolor{okGray}{34.51} & \textcolor{okGray}{36.03} & \textcolor{okGray}{9.17} & \textcolor{okGray}{306.3} & \textcolor{okGray}{202.6} & \textcolor{okGray}{385} & \textcolor{okGray}{2.01}\\
\midrule
\multirow{3}{*}{8B}
 & \textbf{\Tetra{}} & \textbf{2.75} & \textbf{9.71} & \textbf{69.58} & \textbf{88.63} & \textbf{144.5} & \textbf{109.6} & \textbf{407} & \textbf{3.42}\\
 & IQ2\_S & 2.80 & 10.75 & 65.45 & 85.60 & 208.4 & 154.2 & 597 & 3.35\\
 & \textcolor{okGray}{UD-IQ2\_XXS} & \textcolor{okGray}{2.54} & \textcolor{okGray}{13.44} & \textcolor{okGray}{60.96} & \textcolor{okGray}{70.51} & \textcolor{okGray}{222.6} & \textcolor{okGray}{162.3} & \textcolor{okGray}{580} & \textcolor{okGray}{3.09}\\
\midrule
\multirow{3}{*}{14B}
 & \textbf{\Tetra{}} & \textbf{2.76} & \textbf{8.44} & \textbf{75.66} & \textbf{92.04} & \textbf{90.0} & \textbf{71.7} & \textbf{458} & \textbf{5.71}\\
 & IQ2\_S & 2.69 & 9.47 & 72.77 & 91.89 & 123.5 & 99.7 & 613 & 5.40\\
 & \textcolor{okGray}{UD-IQ2\_XXS} & \textcolor{okGray}{2.42} & \textcolor{okGray}{10.47} & \textcolor{okGray}{67.78} & \textcolor{okGray}{85.52} & \textcolor{okGray}{134.0} & \textcolor{okGray}{106.2} & \textcolor{okGray}{599} & \textcolor{okGray}{4.91}\\
\bottomrule
\end{tabular}
\end{table}

\begin{table}[t]
\centering\small\setlength{\tabcolsep}{6pt}
\caption{Our files against IQ2\_S, paired, with 95\,\% intervals: how much
lower our perplexity reads, and how many points higher our files score. Data:
paper2-duel-gaps.csv.}
\label{tab:duelgaps}
\begin{tabular}{llll}
\toprule
 & perplexity, lower by & MMLU, points above & GSM8K, points above\\
\midrule
4B & 25.0\,\% $[23.0, 26.9]$ & 9.08 $[8.19, 9.97]$ & 14.25 $[11.65, 16.86]$\\
8B & 9.7\,\% $[8.3, 11.1]$ & 4.13 $[3.37, 4.88]$ & 3.03 $[1.10, 4.96]$\\
14B & 10.8\,\% $[9.5, 12.2]$ & 2.89 $[2.26, 3.52]$ & 0.15 $[-1.27, 1.58]$, not separated\\
\bottomrule
\end{tabular}
\end{table}

IQ2\_S is what a llama.cpp user would pick at our memory, so we compared it
with our files on the same card, the same day, under one preregistration
(Table~\ref{tab:duel}, Figure~\ref{fig:frontier}). The result is a trade: our
files score higher, and llama.cpp runs faster. Bits per parameter are counted
here from file bytes on both sides. That puts our files 0.03 to 0.09 above
the served count of Table~\ref{tab:main}, because the file also carries the
tokenizer and stores the tail and the row scales in wider types
(Table~\ref{tab:files}).

\paragraph{Quality.} Our files win perplexity and MMLU at every size, by a
margin that shrinks as the model grows (Table~\ref{tab:duelgaps}). The MMLU
lead goes from 9.08 points at 4B to 2.89 at 14B. On GSM8K our files lead at 4B
and 8B, but at 14B the two are not separated, where we had predicted a win.
Two caveats apply. The 14B IQ2\_S file is 0.07 bits per parameter smaller than
ours. And our scores were not measured again that day: they are those of
Tables~\ref{tab:main} and~\ref{tab:gsm8k}.

\paragraph{Speed.} llama.cpp decodes faster at every size and both prompt
lengths. After a 5-token prompt it leads by 60.1, 63.9 and 33.5 tokens per
second at 4B, 8B and 14B, and after 8{,}192 tokens by 34.6, 44.6 and 28.0.
Because the engines differ we form no ratio, but the order is the same in
bytes of weights read per second (Table~\ref{tab:duel}). The speeds of that
table are the decode alone, so ours sit 1.4 to 4.3\,\% above those of
Table~\ref{tab:main}, which include the prompt.

\paragraph{Memory after a short prompt.} After a short prompt the two hold
about as much memory, once llama.cpp's host memory is counted. The card peaks
lie within about 0.3~GB of each other at every size (Table~\ref{tab:duel}).
llama.cpp also keeps its input embedding in host memory, 0.27, 0.27 and
0.33~GB that the card does not see. With it, IQ2\_S holds 2.43, 3.62 and
5.73~GB: more than our files at 4B and 8B, and as much at 14B. Long prompts
change this picture (\S\ref{sec:context}).

\paragraph{One rung below.} A llama.cpp file one rung below ours in size loses
much more quality. UD-IQ2\_XXS holds 0.2 to 0.3 fewer bits per parameter than
our files and scores 27.34, 8.62 and 7.88 MMLU points below them. At 4B it
loops under greedy decoding: 999 of its 1{,}319 GSM8K answers reach the token
limit, so its score of 9.17 measures the loop.

\subsection{Long prompts and several requests}
\label{sec:context}

\begin{figure}[t]
\centering
\includegraphics{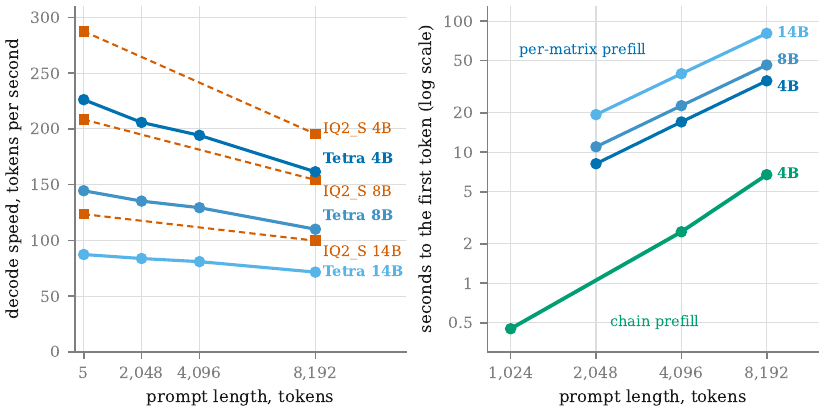}
\caption{Left: decode speed after a prompt of 5 to 8{,}192 tokens, our files
(circles) and IQ2\_S in llama.cpp (squares, measured at the two ends). Right:
seconds from the prompt to the first token. The per-matrix prefill serves the
8B and 14B, the chain prefill the 4B. Data: paper2-context.csv,
paper2-duel.csv.}
\label{fig:context}
\end{figure}

\begin{table}[t]
\centering\small\setlength{\tabcolsep}{6pt}
\caption{Decode speed and time to the first token against prompt length, on
one L40S, as medians of five rounds. An empty cell was not measured.
$^{a}$The two prefills of the 4B timed together in a later job. Data:
paper2-context.csv.}
\label{tab:context}
\begin{tabular}{llrrrrr}
\toprule
 & & \multicolumn{5}{c}{prompt length, tokens}\\
\cmidrule(lr){3-7}
 & & 5 & 1{,}024 & 2{,}048 & 4{,}096 & 8{,}192\\
\midrule
\multirow{3}{*}{decode, tok/s}
 & 4B & 226.2 & & 205.8 & 194.2 & 161.5\\
 & 8B & 144.5 & & 135.2 & 129.3 & 110.0\\
 & 14B & 87.3 & & 83.7 & 80.9 & 71.5\\
\midrule
\multirow{5}{*}{first token, s}
 & 4B, per-matrix prefill & 0.02 & & 8.17 & 17.04 & 35.05\\
 & 8B, per-matrix prefill & 0.03 & & 11.00 & 22.66 & 46.30\\
 & 14B, per-matrix prefill & 0.06 & & 19.36 & 39.72 & 80.74\\
 & 4B, per-matrix prefill$^{a}$ & 0.02 & 3.96 & & 17.79 & 37.22\\
 & \textbf{4B, chain prefill}$^{a}$ & \textbf{0.02} & \textbf{0.45} & & \textbf{2.47} & \textbf{6.74}\\
\bottomrule
\end{tabular}
\end{table}

\begin{figure}[t]
\centering
\includegraphics{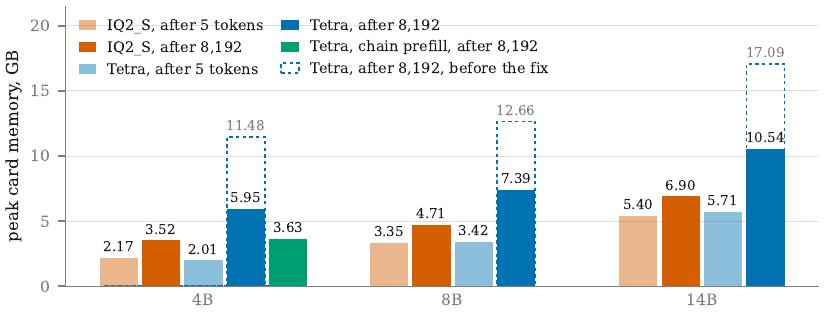}
\caption{Peak card memory after a prompt of 5 and of 8{,}192 tokens. The
dashed outline is our peak on the day of the comparison, before the prefill
kept one score tensor instead of three. Data: paper2-peak.csv.}
\label{fig:peak}
\end{figure}

\paragraph{Decode speed.} A long prompt slows the decode by 18 to 29\,\%. From
a 5-token prompt to one of 8{,}192 tokens, our files go from 226.2 to 161.5
tokens per second at 4B, from 144.5 to 110.0 at 8B and from 87.3 to 71.5 at
14B (Table~\ref{tab:context}, Figure~\ref{fig:context}, left). IQ2\_S stays
ahead at 8{,}192 tokens (Table~\ref{tab:duel}).

\paragraph{Time to the first token.} The per-matrix prefill makes a long
prompt wait, and the chain prefill removes most of that wait at 4B. After
8{,}192 prompt tokens, the per-matrix prefill delivers the first token after
35.05~s at 4B, 46.30~s at 8B and 80.74~s at 14B. Timed with it in one job, the
chain prefill (\S\ref{sec:4e}) brings the 4B from 37.22~s to 6.74~s
(Table~\ref{tab:context}, Figure~\ref{fig:context}, right). The 256 tokens
that follow are the same after either prefill, and on 200 GSM8K problems both
give 165 right answers with no discordant pair. The 8B and the 14B still use
the per-matrix prefill. We did not measure llama.cpp's time to the first
token.

\paragraph{Card memory.} After a long prompt our files hold more card memory
than IQ2\_S (Figure~\ref{fig:peak}). On the day of the comparison they peaked
at about three times what we had predicted, because the prefill's attention
kept three score tensors alive at once. Keeping one, with the same logits bit
for bit, brings the peak after 8{,}192 tokens to 5.95, 7.39 and 10.54~GB.
IQ2\_S peaks at 3.52, 4.71 and 6.90~GB, still 2.43, 2.68 and 3.64~GB lower. At
4B the chain prefill peaks at 3.63~GB, read in its own job.

\paragraph{Several requests.} With concurrent requests our engine falls behind
AWQ in vLLM. A batched form of the kernel decodes requests that arrive
together, while a lone request stays on the persistent kernel. From 1 to 128
concurrent requests, the 4B goes from 221.9 to 3{,}067.5 tokens per second in
aggregate, and every answer equals its batch-1 answer. AWQ in vLLM goes from
203.0 to 10{,}146.2. The batched engine also allocates its KV caches at load,
19.3~GB at 4B for 64 slots of 2{,}048 positions, and it is not the default.

\subsection{One size up for about as much memory}
\label{sec:sizeup}

\begin{table}[t]
\centering\small\setlength{\tabcolsep}{4pt}
\caption{Our 8B and 14B files against the models one size below, and our 14B
against FP16's 4B: the first model minus the second, in points, paired, with
95\,\% intervals. The GSM8K pairs cross the two engines of
\S\ref{sec:gsm8k}. The last two rows show
what one size is worth to AWQ. Weights as in Table~\ref{tab:main}. Data:
paper2-sizeup.csv.}
\label{tab:sizeup}
\begin{tabular}{llrll}
\toprule
\Tetra{} & against & weights GB & MMLU & GSM8K\\
\midrule
8B & AWQ 4B & 2.76 vs 2.67 & $+1.44$ $[+0.71, +2.19]$ & $-0.38$ $[-2.14, +1.39]$\\
14B & AWQ 8B & 5.04 vs 6.10 & $+1.87$ $[+1.20, +2.57]$ & $-0.91$ $[-2.35, +0.53]$\\
8B & FP16 4B & 2.76 vs 8.04 & $-0.56$ $[-1.27, +0.18]$ & $-3.49$ $[-5.18, -1.79]$\\
14B & FP16 8B & 5.04 vs 16.38 & $+0.61$ $[-0.06, +1.27]$ & $-1.21$ $[-2.64, +0.21]$\\
14B & FP16 4B & 5.04 vs 8.04 & $+5.52$ $[+4.79, +6.24]$ & $-0.08$ $[-1.54, +1.39]$\\
\midrule
\multicolumn{2}{l}{AWQ 8B against AWQ 4B} & 6.10 vs 2.67 & $+5.65$ $[+4.96, +6.37]$ & $+3.94$ $[+2.43, +5.46]$\\
\multicolumn{2}{l}{AWQ 14B against AWQ 8B} & 9.98 vs 6.10 & $+4.34$ $[+3.67, +5.00]$ & $+2.43$ $[+1.23, +3.63]$\\
\bottomrule
\end{tabular}
\end{table}

Our aim is to fit a larger model on the same hardware (\S\ref{sec:intro}), so
the fair opponent of our 8B is a 4B that takes about as much memory.
Table~\ref{tab:sizeup} compares our 8B and 14B files with the models one size
below. The three Qwen3 models read the same prompt tokens, so the scores of
Tables~\ref{tab:main} and~\ref{tab:gsm8k} pair question by question, and no
model ran for this section. We chose this comparison after measuring, so it
was not preregistered.

At about the same memory, our larger file beats AWQ's smaller one on MMLU and
is not separated from it on GSM8K. Our 8B and 14B score above AWQ's 4B and 8B
on MMLU (both $p<10^{-3}$), with 0.10~GB more and 1.05~GB less weights, from
unrounded bytes. That gain is 25 and 43\,\% of what one size is worth to AWQ.
On GSM8K the same pairs are not separated ($p=0.74$ and $0.26$), so the gain
of the larger model does not show there.

Against FP16 one size below, our files hold on MMLU and less well on GSM8K.
They are not separated from it on MMLU. On GSM8K our 8B loses to FP16's 4B,
and our 14B is not separated from FP16's 8B. Our 14B also scores above FP16's
4B on MMLU with less memory, and is not separated from it on GSM8K. We compare
no speed here, since the engines differ. AWQ's 4B in vLLM decodes faster than
our 8B in ours (Table~\ref{tab:main}).

\paragraph{Where the larger model fits.} With the KV cache counted, there is
a range of memory budgets where our files run a larger model than AWQ. At
8{,}192 tokens in f16, the KV
cache takes 1.21~GB for Qwen3-4B and 8B, which share its shape, and 1.34~GB
for the 14B. Weights plus that cache take 2.59, 3.97 and 6.39~GB for our 4B,
8B and 14B, and 3.87, 7.31 and 11.32~GB for AWQ. These budgets are computed
from unrounded bytes and leave out the CUDA context and the activations. From
3.97 to 11.32~GB, our files run a larger model than AWQ: one size up, or two
between 6.39 and 7.31~GB, where our 14B scores 7.52 MMLU and 3.03 GSM8K points
above AWQ's 4B. From 11.32~GB both run the 14B, the largest we serve. With
4-bit embedding tables like ours, AWQ's budgets would be 3.31, 5.52 and
9.08~GB (computed; its quality with those tables is not measured), and the
windows would narrow to 3.97 to 5.52~GB for our 8B and 6.39 to 9.08~GB for our
14B. Measured on the card after an 8{,}192-token prompt, our files peak
above these computed budgets, at 5.95, 7.39 and 10.54~GB
(\S\ref{sec:context}). We have no such peak for AWQ.

\subsection{Where the gain comes from}
\label{sec:chain}

\begin{table}[t]
\centering\small\setlength{\tabcolsep}{5pt}
\caption{How each file is built, one step per row, each gain paired against the
row above on the 14{,}042 questions. Bits per parameter count the file as
served, with 8-bit embedding tables in the first two rows of each size. Those
rows were scored with f16 tables, so the last gain also includes the cost of
4-bit tables. Data: paper2-chain.csv.}
\label{tab:chain}
\begin{tabular}{llrrrl}
\toprule
 & step & b/param & MMLU & gain & 95\,\% interval, McNemar\\
\midrule
\multirow{3}{*}{4B}
 & \Tetra{} $+$ int4 \texttt{v\_proj} & 2.7475 & 57.95 & & \\
 & $+$ trained row scales & 2.7475 & 61.11 & $+3.15$ & $[+2.56, +3.74]$, $p=1.8\cdot10^{-25}$\\
 & $+$ int4 $o$, down; 4-bit tables & 2.7320 & 63.37 & $+2.26$ & $[+1.66, +2.87]$, $p=8.2\cdot10^{-13}$\\
\midrule
\multirow{3}{*}{8B}
 & \Tetra{} $+$ int4 \texttt{v\_proj} & 3.0683 & 64.87 & & \\
 & $+$ trained row scales & 3.0683 & 68.16 & $+3.29$ & $[+2.78, +3.80]$, $p=7.2\cdot10^{-38}$\\
 & $+$ int4 down; 4-bit tables & 2.6953 & 69.58 & $+1.42$ & $[+0.93, +1.92]$, $p=3.2\cdot10^{-8}$\\
\midrule
\multirow{3}{*}{14B}
 & \Tetra{} $+$ int4 \texttt{v\_proj} & 2.7371 & 72.53 & & \\
 & $+$ trained row scales & 2.7371 & 74.20 & $+1.67$ & $[+1.27, +2.08]$, $p=5.0\cdot10^{-16}$\\
 & $+$ int4 $o$, down; 4-bit tables & 2.7305 & 75.66 & $+1.46$ & $[+0.99, +1.94]$, $p=1.4\cdot10^{-9}$\\
\bottomrule
\end{tabular}
\end{table}

Both steps of the recipe gain MMLU at every size, and all six gains are
separated from zero (Table~\ref{tab:chain}).

\paragraph{Perplexity does not follow the last step.} The last step, int4
projections and 4-bit tables, raises the perplexity of the 4B and the 8B, on
the windows of \S\ref{sec:setup}. It goes from 12.33
to 12.58 at 4B, $+2.1\,\%$ $[1.5, 2.6]$, and from 9.57 to 9.71 at 8B,
$+1.4\,\%$ $[1.1, 1.7]$, worse on all 12 windows at both sizes. At 14B it goes
from 8.46 to 8.44, $-0.2\,\%$ $[-0.9, 0.5]$, not separated. So the step gains
MMLU at every size and costs perplexity at the two smaller ones. We did not
measure which half of the step costs perplexity, the 4-bit tables or the
projections moved to int4. Data: paper2-ppl-sealing.csv.

\paragraph{Which projections get the int4 bytes.} At 8B the embedding tables
take the largest share of the parameters (\S\ref{sec:q4emb}), so at about 2.7
bits per parameter int4 gets fewer projections than at 4B and 14B
(Table~\ref{tab:files}). Our 8B file, with int4 on \texttt{down\_proj} of
layers 10 to 26 and not on \texttt{o\_proj}, scores 0.77 points $[0.26, 1.28]$
above int4 on all of \texttt{o\_proj} plus \texttt{down\_proj} of layers 15
to 20, which scores 68.81 at 2.7047 bits per parameter ($p=0.004$). We had predicted the opposite order. The 4B and
14B files keep \texttt{o\_proj} in int4, and we have not tested the 8B choice
there.

\paragraph{Selection on the test set.} These choices were made on MMLU test
questions, with no held-out split. At 8B we kept the better of the two files
above, under a preregistered rule. At 4B the window of layers 12 to 23 was the
best of three windows of twelve layers, scored on the full test set with an
earlier 4B file. The projection types sent to int4 were ranked at 4B on
samples of 2{,}280 questions. The 14B window follows a rule fixed before
measuring. It is the largest centred window that the bytes freed by the
tables pay for, as at 4B. So the MMLU scores of our files carry a
selection bias that we did not measure. For \texttt{o\_proj},
the one choice checked on questions that did not select it, the gain is
$+3.12$ points on the 2{,}280 that selected it and $+1.55$ on the other
11{,}762. GSM8K and perplexity chose nothing. We also tried a criterion that reads no
benchmark, the ratio of the Hessian-weighted errors of \Tetra{} and int4 per
matrix. On an earlier 4B file it did not rank six of the seven projection types
(\texttt{v\_proj} was already int4) as MMLU does (Spearman $\rho = 0.54$,
$p = 0.30$, an exploratory reading), so it chose nothing here.

\subsection{Where the 4B still loses its points}
\label{sec:ablation}

\begin{figure}[t]
\centering
\includegraphics{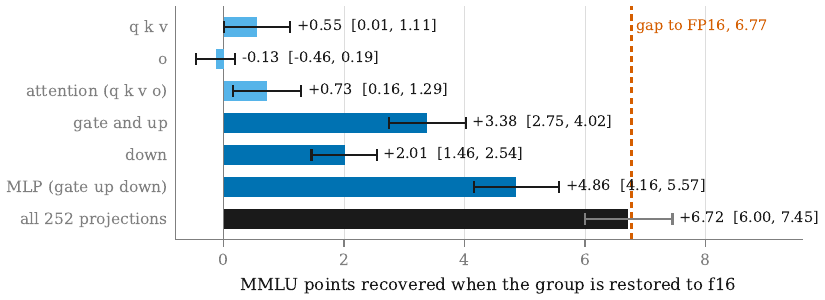}
\caption{MMLU points the sealed 4B recovers when one group of projections is
restored to the f16 weights of the checkpoint, on the 14{,}042 questions, with
95\,\% paired intervals. The dashed line is the file's gap to FP16. Data:
paper2-ablation.csv.}
\label{fig:ablation}
\end{figure}

The MLP carries most of the 4B's gap to FP16. To locate those 6.77 points, we
restored one group of projections at a time to the f16 weights of the
checkpoint and scored the full test set (Figure~\ref{fig:ablation}). Restoring
\texttt{gate\_proj} and \texttt{up\_proj} recovers 3.38 points,
\texttt{down\_proj} 2.01, and the three together 4.86. The attention
projections recover 0.73, and $o$ alone recovers nothing that the test
resolves. With all 252 projections restored, the file recovers 6.72 points and
sits 0.04 $[-0.23, 0.31]$ below FP16, so the 4-bit embedding and head cost
nothing that the test resolves either.

The groups do not add up. All projections together recover 1.14 points
$[0.43, 1.85]$ more than the MLP and the attention taken apart. This reading
covers one calibration draw, MMLU only, on the dense reconstruction.

\subsection{A second architecture}
\label{sec:a100}

\begin{table}[t]
\centering\small\setlength{\tabcolsep}{5pt}
\caption{The kernel benchmark and the served files on one NVIDIA
A100-SXM4-80GB (sm\_80, HBM2e), with the protocols of Table~\ref{tab:bench}
and Table~\ref{tab:main}. All six kernel arms time the same 252 matrices in
one process, at tile 128, and the ratios are formed round by round. Data:
a100-bench.csv, a100-served.csv.}
\label{tab:a100}
\begin{tabular}{lrrrr}
\toprule
 & median ms & \bw{} in VRAM & GB/s, fastest round & against FP16\\
\midrule
FP16 & 6.899 & 16.000 & 1054 & 1.00$\times$\\
FP16, cuBLAS & 6.050 & 16.000 & 1205 & 1.14$\times$\\
AWQ w4g128 & 3.777 & 4.179 & 503 & 1.83$\times$\\
\Planes{} & 8.700 & 4.804 & 251 & 0.79$\times$\\
\textbf{\Tetra{}} & \textbf{6.224} & \textbf{2.150} & \textbf{157} & \textbf{1.11$\times$}\\
no weights & 4.070 & 0.159 & 18 & 1.69$\times$\\
\midrule
\textit{decode, tokens per second} & 4B & 8B & 14B & \\
\Tetra{}, our engine & 74.2 & 57.9 & 39.0 & \\
\Tetra{}, f16 tables, our engine & 45.9 & 34.2 & 23.7 & \\
f16 dense path, our engine & 51.1 & 37.4 & 25.1 & \\
FP16, vLLM & 145.6 & 91.8 & 52.7 & \\
AWQ w4g128, vLLM & 240.2 & 188.5 & 131.7 & \\
\bottomrule
\end{tabular}
\end{table}

On a card with faster memory, the kernel keeps less of its advantage. The
L40S is an Ada GPU with GDDR6 memory at
0.86~TB/s~\citep{nvidia_l40s_datasheet}. To test a card where reading fewer
bytes matters less, we ran the kernel benchmark and the three sealed files on
an A100-SXM4-80GB, an Ampere GPU (sm\_80) with HBM2e at about
2.0~TB/s~\citep{nvidia_a100_datasheet} (Table~\ref{tab:a100}). On this card
\Tetra{} runs faster than our FP16 kernel, where \Planes{} runs slower, but
cuBLAS reads the same f16 weights faster still, and AWQ stays well ahead.

The served files gain nothing from their projections on this card, because
our dense path runs its own through cuBLAS. With f16 tables on both sides, as in
\S\ref{sec:main}, the served files decode at 0.92, 0.91 and 0.95 times the speed of the dense
arm of the same runs (50.1, 37.4 and 25.1 tokens per second), where they ran
faster on the L40S. Their lead with 4-bit tables comes
from the tables alone, and they stay below FP16 in vLLM at every size. Memory
and quality do not depend on the card. The greedy tokens match the dense path for all 256 at 4B
and 8B and first differ at token 78 at 14B, as on the L40S. These decode
speeds are those of the per-matrix path with the expanded attention. The persistent kernel
of \S\ref{sec:4e} has not run on this card.

\section{Limitations}
\label{sec:limitations}

\begin{itemize}
\item \textbf{Two data-center GPUs.} We measured speed on one NVIDIA L40S and
  one A100 (\S\ref{sec:a100}), never on a consumer GPU, the hardware this work
  targets. The persistent kernel ran on the L40S only.
\item \textbf{One calibration draw.} Every size is encoded from one draw of
  calibration text (\S\ref{sec:setup}). At 4B, with the earlier codebook, the
  spread across three draws exceeds every gain of Table~\ref{tab:chain}
  (\S\ref{sec:kept}). Those gains compare fixed files, but our gaps to FP16,
  AWQ and IQ2\_S do not include that spread.
\item \textbf{The scores and the speeds come from two forms of the engine.}
  MMLU and perplexity are scored on the dense reconstruction, and GSM8K on the
  per-matrix path, all with the expanded attention. The served engine is
  tied to them by a perplexity tolerance, by token identity and by a
  50-problem pilot per size (\S\ref{sec:setup}), not by a full run of each
  benchmark.
\item \textbf{Selection on the test set.} Our int4 choices used MMLU test
  questions (\S\ref{sec:chain}).
\item \textbf{GSM8K is an easy test for these models.} FP16 scores 92 to 95,
  and the problems have been public since 2021. We did not test Qwen3's
  thinking mode, which writes much longer chains, nor a harder math
  benchmark.
\item \textbf{Greedy decoding at batch 1.} The persistent kernel serves one
  request at a time and does not sample. Several requests go through a
  batched kernel that AWQ in vLLM outruns at every batch size
  (\S\ref{sec:context}).
\item \textbf{Long prompts.} After 8{,}192 prompt tokens our files hold 2.4 to
  3.6~GB more card memory than IQ2\_S in llama.cpp, and the 8B and 14B wait
  46 and 81~s for their first token (\S\ref{sec:context}).
\item \textbf{One llama.cpp build, one importance matrix.} The IQ2\_S files
  are ours, built from the calibration text of our files. Another build or
  another matrix may score differently.
\end{itemize}

\section{Conclusion}
\label{sec:conclusion}

The GPU can read a Leech-lattice code without expanding it, if the codebook
bounds each third of a block separately. Three more changes, which leave the
lattice codes untouched, make complete Qwen3 files at about 2.7 bits per
parameter. They stay below 4-bit AWQ on MMLU and GSM8K, and perplexity ranks
them differently. At the same bits they score above llama.cpp's IQ2\_S on
MMLU and perplexity at every size, by a margin that shrinks from 9.08 points
at 4B to 2.89 at 14B.

Running a token in one launch makes the 4B 1.72 times faster on the L40S and
leaves its tokens unchanged. llama.cpp still decodes faster and holds less
memory after a long prompt, and the kernel's gain depends on the card: on an
A100, cuBLAS reads f16 weights faster than our kernel reads \Tetra{}.

Next, in this order, we would carry the chain prefill to the 8B and the 14B,
score MMLU and GSM8K in full through the served engine, add a second
calibration draw, and measure a consumer GPU. The ablation of
\S\ref{sec:ablation} places the remaining loss at 4B in the MLP. Qwen3-32B
has not been encoded.

\section*{Availability}

The code, the measurement logs behind every number, the preregistrations and
this paper's source are at \url{https://github.com/pjmalandrino/llvq}. They
are under MIT or Apache-2.0, and the paper under CC BY 4.0. Each number in
this paper is \emph{measured} (read off a run), \emph{computed} (arithmetic on
measured quantities) or \emph{cited}. \mbox{\nolinkurl{paper2/PROVENANCE.md}}
names the source of each, and \mbox{\nolinkurl{paper2/README.md}} gives the
commands that rerun the main measurements. A script draws every data figure
from a CSV in \texttt{docs/data/}, and a second one stops the build if a table
no longer matches its CSV.

The three sealed files are on the Hugging Face Hub, with these SHA-256 digests:
\begin{itemize}
\item \nolinkurl{qwen3-4b-sealed.bin}, \texttt{886391a8}\ldots\\
  \url{https://huggingface.co/Pier-Jean/Qwen3-4B-LLVQ-Tetra-sealed}
\item \nolinkurl{qwen3-8b-sealed-B.bin}, \texttt{7bdb9a55}\ldots\\
  \url{https://huggingface.co/Pier-Jean/Qwen3-8B-LLVQ-Tetra-sealed}
\item \nolinkurl{qwen3-14b-sealed.bin}, \texttt{61db37fe}\ldots\\
  \url{https://huggingface.co/Pier-Jean/Qwen3-14B-LLVQ-Tetra-sealed}
\end{itemize}

Each file is served under a configuration file in \texttt{configs/}, named
after its size and ending in \texttt{-tetra-e4-4e-v9.json}. The 4B's chain
prefill is \nolinkurl{qwen3-4b-tetra-e4-4e-v9-chain.json}.

The QTIP kernel of Appendix~\ref{app:bench} is under GPL~v3, so we do not
redistribute it. The benchmark downloads it at a fixed commit.

\section*{Use of generative AI}

Anthropic's Claude, through the Claude Code command-line tool (2026), drafted
and revised the text and drew the figures. It wrote parts of the CUDA kernels,
the Rust host code and the tests. It recomputed statistics from the committed
measurement logs. The author designed the study, authorized every
measurement, checked each number against its source, and is responsible for
the whole paper.

\bibliographystyle{plainnat}
\bibliography{refs}

\appendix
\needspace{24\baselineskip}
\section{The ten-arm kernel benchmark}
\label{app:bench}

\begin{table}[H]
\centering\small\setlength{\tabcolsep}{5pt}
\caption{Each row is one kernel timed on the 252 matrices of the 4B on one L40S.
The first ten rows are the five arms of Table~\ref{tab:bench} and five more,
timed in the same process at tile 64 with the same protocol. The rows marked
``earlier'' are the layouts of \citet{llvq1preprint}, timed again here.
$^{*}$The QTIP measurement of \citet{llvq1preprint}, from another process
(2026-08-21): the HYB kernel of QTIP's repository ($L=16$, $Q=9$, $V=2$) on
pseudo-random codes, in its own launch geometry. It reads no row scales and no
tail, and its 2~KiB table is not counted. It runs below the no-weights
control. Data: echelle-formats.csv.}
\begin{tabular}{lrrrr}
\toprule
arm & median ms & GB read & \bw{} & GB/s, fastest round\\
\midrule
FP16, our control & 11.022 & 7.27 & 16.000 & 660\\
FP16, cuBLAS & 10.839 & 7.27 & 16.000 & 671\\
\textsc{Slot32} (ours, earlier) & 5.737 & 2.50 & 5.510 & 437\\
\Planes{} (ours, earlier) & 5.004 & 2.18 & 4.804 & 436\\
\textsc{Planes12x} (ours, earlier) & 5.385 & 1.97 & 4.342 & 366\\
\textsc{Golay70} (ours, earlier) & 8.124 & 1.63 & 3.589 & 201\\
\textsc{Golay70}, hoisted (ours, earlier) & 6.060 & 1.63 & 3.589 & 269\\
\Tetra{} (this paper) & 3.734 & 0.98 & 2.150 & 262\\
AWQ w4g128~\citep{awq2024} & 3.263 & 1.90 & 4.179 & 582\\
no-weights control & 2.294 & 0.07 & 0.159 & 32\\
QTIP 2-bit, HYB~\citep{qtip2024}$^{*}$ & 2.246 & 0.91 & 2.000 & 405\\
\bottomrule
\end{tabular}
\end{table}

\section{Record of predictions}
\label{app:record}

\paragraph{Predictions and how they scored.} From 2026-08-10 on, before each
measurement campaign, we wrote a timestamped preregistration naming the arms
(the configurations compared), the decision rules and a prediction. These
files are never edited, and a departure from the plan is written next to its
file. The FP16 and AWQ perplexities come from earlier runs, with no
preregistration. Five later results have none either: the retention figures of
\S\ref{sec:kept}, measured on the development machine; the comparison of
\S\ref{sec:sizeup}, chosen after measuring; the Hessian reading of
\S\ref{sec:chain}, which is exploratory; the paired cost of sealing at 14B in
\S\ref{sec:chain}, a reading chosen after the run; and the IQ2\_XXS speed of
Table~\ref{tab:main}. The comparison with IQ2\_S, the persistent kernel's
speeds, the peak memory, the chain prefill and the ablation of
\S\ref{sec:ablation} each have a preregistration. Table~\ref{tab:predictions} lists a selection of the
predictions; the preregistrations in \texttt{proofs/} hold all of them. A
signed value with no unit is in MMLU points, except in the GSM8K rows. Rows
that list several values give the point predictions; their intervals are in
the preregistrations, and every listed value fell inside its interval.

Several campaigns departed from their preregistration, and two of them
overrode a decision rule. The 14B row-scale training ended with its loss gauge
reading ``not improved'', where its preregistration said to stop. We folded
the scales anyway, on a decision recorded before the result was known. That
step is the $+1.67$ of Table~\ref{tab:chain}. The perplexity campaign said
that a sealed file below its trained base voids the run, and the 14B read
8.4440 against 8.4622. We published it anyway, after the run, because the
paired interval contains zero (\S\ref{sec:chain}). The chain prefill was
served although its preregistration read ``not settled'' on one bound: the
decode after it is 2.9 and 3.2\,\% faster than after the per-matrix prefill,
outside a two-sided band. The comparison with IQ2\_S missed our peak memory
after a long prompt by a factor of 2.3 to 3.2, which led to the fix of
\S\ref{sec:context}. In the ablation, five of eight signed predictions fell
inside their intervals, and the attention group cost less than predicted. The
other departures changed how a run was launched, not what it measured.

\begin{table}[!t]
\centering\small\setlength{\tabcolsep}{3pt}
\caption{Preregistered predictions against the measurements. Sources: the
preregistrations in \texttt{proofs/} and the logs in \texttt{docs/mesures/}.}
\label{tab:predictions}
\begin{tabular}{lrrl}
\toprule
quantity & predicted & measured & inside?\\
\midrule
\Tetra{} time, 252 matrices, tile 64 & 3.72~ms $[3.60, 3.85]$ & 3.734~ms & yes\\
\Tetra{} ms at tiles 128, 64, 32 & 4.36, 3.72, 3.86 & 4.35, 3.70, 3.84 & yes, all three\\
row-scale training, 4B & $+3.5$ $[+1.5, +4.5]$ & $+3.15$ & yes\\
row-scale training, 8B & $+2.0$ $[+0.5, +3.5]$ & $+3.29$ & yes\\
row-scale training, 14B & $+3.2$ $[+1.7, +4.7]$ & $+1.67$ & no, 0.03 under\\
sealed composition, 4B & $+1.2$ $[+0.2, +2.2]$ & $+2.26$ & no, 0.06 over\\
sealed composition, 8B, down only & $+0.7$ $[-0.3, +1.7]$ & $+1.42$ & yes\\
sealed composition, 14B & $+1.0$ $[0, +2.0]$ & $+1.46$ & yes\\
8B at 2.7, $o$ and down minus down only & $+0.2$ $[-0.6, +1.0]$ & $-0.77$ & no, 0.17 under\\
second training (row scales and RMSNorm), 4B & $+0.8$ $[-0.4, +2.0]$ & $+0.02$ & yes\\
AWQ 4B, full test set & 69.9 $[69.2, 70.6]$ & 68.14 & no, 1.06 under\\
IQ2\_XXS 4B, full test set & 39.0 $[37.5, 40.5]$ & 39.78 & yes\\
FP16 8B, full test set & 77.0 $[75.5, 78.5]$ & 75.05 & no, 0.45 under\\
14B, \Tetra{} $+$ int4 \texttt{v\_proj}, full test set & 68.3 $[65.3, 71.2]$ & 72.53 & no, 1.33 over\\
AWQ tok/s in vLLM, 8B & 90 $[75, 110]$ & 123.2 & no, 13.2 over\\
AWQ tok/s in vLLM, 14B & 55 $[45, 70]$ & 77.8 & no, 7.8 over\\
4B tok/s, GB, per-matrix path & 110 $[95, 125]$, 1.38 & 113.8, 1.38 & yes\\
8B tok/s, GB, per-matrix path & 100 $[88, 112]$, 2.77 & 95.0, 2.76 & yes\\
14B tok/s, GB, per-matrix path & 58 $[50, 66]$, 5.04 & 57.2, 5.04 & yes\\
4B tok/s, persistent kernel & 220 $[210, 230]$ & 221.5 & yes\\
8B tok/s, persistent kernel & 140 $[110, 175]$ & 142.5 & yes\\
14B tok/s, persistent kernel & 85 $[68, 105]$ & 86.3 & yes\\
persistent over per-matrix, 4B & $1.75\times$ $[1.65, 1.85]$ & $1.717\times$ & yes\\
tok/s after 8{,}192 tokens, three sizes & 160.0, 100, 62 & 161.5, 110.0, 71.5 & yes, all three\\
IQ2\_S perplexity, 4B, 8B, 14B & 18, 12, 9.8 & 16.78, 10.75, 9.47 & yes, all three\\
IQ2\_S MMLU, 4B, 8B, 14B & 50, 60, 68 & 54.29, 65.45, 72.77 & no at 14B, above\\
IQ2\_S GSM8K, 4B, 8B, 14B & 60, 75, 85 & 68.23, 85.60, 91.89 & above at 8B, 14B\\
\Tetra{} above IQ2\_S on GSM8K, 14B & a win & not separated & no\\
\Tetra{} peak GB after 8{,}192 tokens & 3.6, 4.9, 7.3 & 11.48, 12.66, 17.09 & no, all over\\
256 identical tokens & at each size & 4B, 8B; 14B to 78 & no at 14B\\
GSM8K, FP16 at 4B, 8B, 14B & 90, 92, 94 & 92.12, 93.25, 95.30 & yes, all three\\
GSM8K, AWQ at 4B, 8B, 14B & 88, 91, 93 & 89.01, 92.95, 95.38 & yes, all three\\
GSM8K, \Tetra{} 4B & 75 $[67, 83]$ & 82.49 & yes\\
GSM8K, \Tetra{} 8B & 82 $[75, 88]$ & 88.63 & no, 0.63 over\\
GSM8K, \Tetra{} 14B & 88 $[83, 92]$ & 92.04 & no, 0.04 over\\
GSM8K, \Tetra{} minus FP16, 4B & $-15$ $[-23, -8]$ & $-9.63$ & yes\\
GSM8K, \Tetra{} minus FP16, 8B & $-10$ $[-17, -5]$ & $-4.62$ & no, 0.38 over\\
GSM8K, \Tetra{} minus FP16, 14B & $-6$ $[-11, -2]$ & $-3.26$ & yes\\
perplexity, sealed 4B, 8B, 14B & 12.51, 9.72, 8.59 & 12.58, 9.71, 8.44 & yes, all three\\
A100, \Tetra{} kernel time & 7.5~ms $[6.5, 9.0]$ & 6.224~ms & no, 0.28~ms under\\
A100, sealed files tok/s & 75, 60, 37 & 74.2, 57.9, 39.0 & yes, all three\\
\bottomrule
\end{tabular}
\end{table}

\end{document}